%% file: main.tex
\documentclass{article}

\usepackage{PRIMEarxiv}
\usepackage{subfigure}

\usepackage[utf8]{inputenc} % allow utf-8 input
\usepackage[T1]{fontenc}    % use 8-bit T1 fonts
\usepackage{hyperref}       % hyperlinks
\usepackage{url}            % simple URL typesetting
\usepackage{booktabs}       % professional-quality tables
\usepackage{amsfonts}       % blackboard math symbols
\usepackage{nicefrac}       % compact symbols for 1/2, etc.
\usepackage{microtype}      % microtypography
\usepackage{lipsum}
\usepackage{fancyhdr}       % header
\usepackage{graphicx}       % graphics
\usepackage{makecell}
\usepackage{float}
\graphicspath{{media/}}     % organize your images and other figures under media/ folder

\input{math_commands}

\theoremstyle{definition}

\theoremstyle{remark}

\title{Multimodal Scientific Learning Beyond\\ Diffusions and Flows
}

\author{
  Leonardo Ferreira Guilhoto \\
  Graduate Group in Applied Mathematics and Computational Science\\
  University of Pennsylvania \\
  Philadelphia, PA, USA\\
  \texttt{guilhoto@sas.upenn.edu} \\
   \And
  Akshat Kaushal \\
  Department of Computer and Information Science\\
  University of Pennsylvania \\
  Philadelphia, PA, USA\\
  \texttt{akaush@seas.upenn.edu} \\
   \And
  Paris Perdikaris \\
  Department of Mechanical Engineering and Applied Mechanics\\
  University of Pennsylvania \\
  Philadelphia, PA, USA\\
  \texttt{pgp@seas.upenn.edu} \\
}

\begin{document}
\maketitle

\input{abstract}

% keywords can be removed

\input{Text/main_body}

\section*{Acknowledgments}
We would like to acknowledge support from the US Department of Energy under the Advanced Scientific Computing Research program (grant DE-SC0024563) and the US National Science Foundation (NSF) Soft AE Research Traineeship (NRT) Program (NSF grant 2152205). We also thank the developers of software that enabled this research, including JAX \cite{jax2018github}, Flax \cite{flax2020github} Matplotlib \cite{Hunter2007matplotlib} and NumPy \cite{harris2020numpy}.

\section*{Code \& Data Availability}
We release our entire codebase publicly as the \texttt{JaxMix} package, available at \url{https://github.com/PredictiveIntelligenceLab/JaxMix}. The \texttt{examples} folder within this repository contains all the experiments used in this paper, including the generation of figures. We hope other computational scientists find this repository useful for experimenting with multimodal modeling across their own problems.

\section*{Impact Statement}
Our contributions enable advances in deep learning and uncertainty quantification. This has the potential to impact a wide range of downstream applications. While we do not anticipate specific negative impacts from this work, as with any powerful predictive tool, there is potential for misuse. We encourage the research community to consider the ethical implications and potential dual-use scenarios when applying these technologies in sensitive domains and to avoid its application altogether to weaponry and other military technologies.

\section*{Author Contributions}
L.F.G. and P.P. conceived the methodology and conceptualized the experiments. L.F.G. and A.K. conducted experiments and analyzed results. P.P. provided funding and supervised this study. All authors helped in writing and reviewing the manuscript.

%Bibliography
\bibliographystyle{unsrt}
\bibliography{references}

\input{Text/appendix}

\end{document}

%% file: math_commands.tex
\usepackage{amsmath,amsfonts,bm}

\def\eqref#1{equation~\ref{#1}}
\def\1{\bm{1}}

\def\vx{{\bm{x}}}

\def\mI{{\bm{I}}}

\DeclareMathAlphabet{\mathsfit}{\encodingdefault}{\sfdefault}{m}{sl}
\SetMathAlphabet{\mathsfit}{bold}{\encodingdefault}{\sfdefault}{bx}{n}

\newcommand{\R}{\mathbb{R}}

\usepackage{amsthm}

%% file: abstract.tex
\begin{abstract}
Scientific machine learning (SciML) increasingly requires models that capture multimodal conditional uncertainty arising from ill-posed inverse problems, multistability, and chaotic dynamics. While recent work has favored highly expressive implicit generative models such as diffusion and flow-based methods, these approaches are often data-hungry, computationally costly, and misaligned with the structured solution spaces frequently found in scientific problems. We demonstrate that Mixture Density Networks (MDNs) provide a principled yet largely overlooked alternative for multimodal uncertainty quantification in SciML. As explicit parametric density estimators, MDNs impose an inductive bias tailored to low-dimensional, multimodal physics, enabling direct global allocation of probability mass across distinct solution branches. This structure delivers strong data efficiency, allowing reliable recovery of separated modes in regimes where scientific data is scarce. We formalize these insights through a unified probabilistic framework contrasting explicit and implicit distribution networks, and demonstrate empirically that MDNs achieve superior generalization, interpretability, and sample efficiency across a range of inverse, multistable, and chaotic scientific regression tasks.
\end{abstract}

%% file: Text/main_body.tex
\section{Introduction}
The integration of deep learning into the natural sciences, often called Scientific Machine Learning (SciML) or AI for Science, has shifted scientific modeling away from purely deterministic surrogates toward probabilistic formulations. In classical scientific computing, uncertainty is often treated as nuisance noise, typically assumed additive and Gaussian, around a single underlying truth. While this unimodal assumption is reasonable for well-posed forward problems, it breaks down in many frontier applications, including chaotic fluid dynamics \cite{mikhaeil2022difficulty} and inverse problems \cite{li2023seismic-inversion, song2022solvinginverseproblemsmedical}. In such settings, the relationship between inputs $x$ and outputs $y$ is frequently non-invertible or ill-posed. Instead, governing physics can admit multiple physically valid outcomes, implying that a given set of control parameters or initial conditions $x$ induces a multimodal conditional distribution over outputs $y$. Reliable models for scientific discovery must therefore represent not just a single point prediction, but the structure of the conditional distribution $p(y\lvert x)$, including separated, physically meaningful modes.

%In magnetic confinement fusion, for example, a predictive model must distinguish between a stable plasma trajectory and a disruptive instability \cite{fasoli2016computationalfusion}; averaging these outcomes yields a non-physical state that is meaningless for control. An analogous issue arises in medical imaging inversion, where multiple anatomically distinct reconstructions may be consistent with the same sensor measurements \cite{song2022solvinginverseproblemsmedical}. In such cases, unimodal approximations that interpolate between plausible solutions obscure clinically or scientifically critical structures.

Despite these requirements, Uncertainty Quantification (UQ) in SciML is still dominated by unimodal modeling assumptions. Most regression models are trained by minimizing Mean Squared Error (MSE), an objective that strictly recovers the conditional expectation $\mathbb{E}[y \lvert x]$. When the true conditional distribution is multimodal, this average can lie in regions of low probability density. As a consequence, the resulting prediction may be physically invalid. More expressive approaches such as Bayesian Neural Networks (BNNs) are, in principle, capable of representing complex posteriors, but in practice often revert to unimodal  approximations due to the limitations of mean-field variational inference and the centralizing effects of standard ensemble aggregation \cite{Izmailov2021WhatAreBNNPosteriors, DeepEnsembles2021}.

In response to the limitations of unimodal UQ, recent work in machine learning has favored highly expressive implicit generative models, including variational autoencoders \cite{kingma2013auto}, denoising diffusion probabilistic models (DDPMs) \cite{ho2020ddpm} and conditional flow matching (CFM) \cite{lipman2022flow}. These methods have achieved remarkable success in high-dimensional generative tasks by learning to transform simple base measures into complex target distributions through iterative denoising or continuous-time dynamics. Their application to scientific regression problems, however, introduces challenges that are not typically encountered in mainstream generative modeling. Recent theoretical analyses \cite{Koehler2023Statistical, Biroli2024Dynamical} indicate that diffusion and flow-based models struggle to accurately learn distributions with well-separated modes, requiring exponentially many samples to correctly capture the relative weights of disconnected solution branches. In addition, the high computational cost of inference, often involving hundreds of model evaluations per prediction, limits their applicability in real-time or resource-constrained scientific settings. These limitations are amplified in the low-data regimes typical of SciML, where training sets are orders of magnitude smaller than in mainstream generative modeling. Diffusion and flow-based models tend to degrade sharply under such constraints, exhibiting poor mode coverage and unstable training.

% These limitations are further amplified in the low-data regimes typical of many SciML applications. Experimental measurements and high-fidelity simulations are often expensive, time-consuming, or physically constrained, resulting in training sets that are several orders of magnitude smaller than those used in mainstream generative modeling such as image generation. Diffusion and flow-based models, whose performance relies on large datasets to accurately learn complex transport maps, tend to degrade sharply under such constraints, exhibiting poor mode coverage and unstable training. This data inefficiency poses a significant barrier to their adoption in scientific settings where reliable uncertainty estimates must be obtained from limited observations.

In light of these observations, this work revisits Mixture Density Networks (MDNs) \cite{bishop1994mixture} as a principled alternative for multimodal uncertainty quantification in SciML. Although MDNs are a well-established class of models in the machine learning literature, they have seen little adoption in modern SciML pipelines, and their behavior under multimodal, low-data physical regimes has not been systematically examined until now. Building on recent theory, we argue that the explicit mixture structure of MDNs induces an inductive bias inherently compatible with disconnected solution sets, allowing them to represent multimodal conditional densities without the topological distortions introduced by continuous transport-based models. We validate these claims through experiments on inverse problems, multistable dynamical systems, and chaotic time-series prediction, demonstrating that MDNs achieve reliable mode recovery and competitive density estimation performance with substantially lower data and computational requirements than diffusion and flow-based approaches.

The main contributions of this paper are as follows:
\begin{itemize}
\item We introduce a {\it unified probabilistic framework} for SciML that situates point-estimate models, explicit density estimators, and implicit generative approaches within a common formalism.
\item We provide a systematic comparison grounded in statistical learning theory  between Mixture Density Networks and non-parametric 
models in low-to-medium dimensional problems, demonstrating that parametric mixture models achieve more reliable mode recovery and data efficiency in regimes typical of SciML.
\item We emphasize the role of {\it interpretability} in scientific modeling by showing how MDN mixture components and weights admit direct physical interpretation, enabling connections to phase diagrams, stability boundaries, and regime classification that are not readily accessible in diffusion or flow-based models.
\item We present numerical experiments across representative SciML benchmarks demonstrating that MDNs achieve parity or superior conditional density estimation performance relative to modern diffusion-based approaches  when operating in the low-data regime.
\item We release \texttt{JaxMix}, a Python package based on JAX to allow computational scientists to explore MDNs and other multimodal modeling approaches. It is publicly available at \url{https://github.com/PredictiveIntelligenceLab/JaxMix}.
\end{itemize}

\section{Multimodal Uncertainty in Scientific Problems}

Scientific machine learning increasingly operates in regimes where deterministic prediction is neither possible nor meaningful. Many physical systems are noisy, partially observed, chaotic, or under-specified, so that a single input admits multiple valid outcomes. Yet, most deep learning methods remain rooted in point estimation via minimizing Mean Squared Error (MSE), implicitly treating variability as noise rather than genuine physical ambiguity. This paradigm produces predictions that can be unphysical or misleading when the true conditional distribution is multimodal.

\begin{figure*}[ht]
    \centering
    \includegraphics[width=0.99\textwidth]{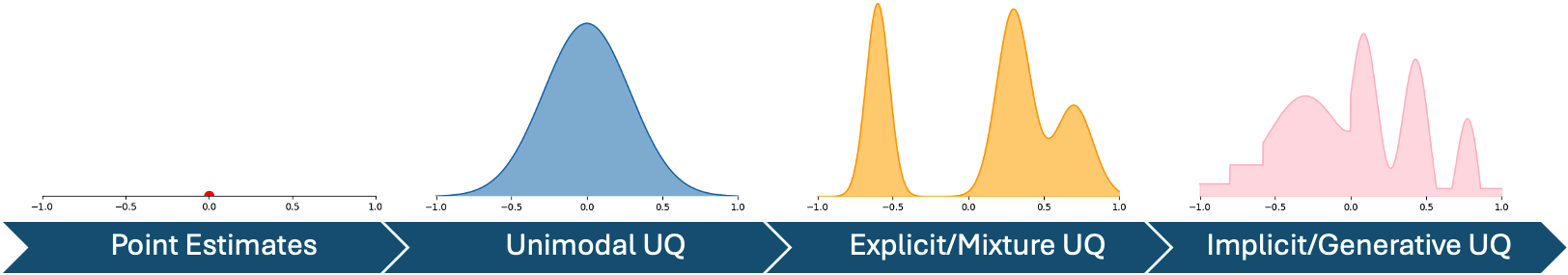}
    \caption{Illustration of increasing UQ complexity. Point estimates provide no uncertainty (left), while arbitrary generative models are powerful but data/compute intensive (right). Many scientific problems are well approximated by simpler multimodal mixture distributions.}
    \label{fig:uq-complexity-spectrum}
\end{figure*}

Figure~\ref{fig:uq-complexity-spectrum} illustrates UQ approaches across a spectrum. At one extreme, point estimates ignore uncertainty entirely. At the other, non-parametric generative models can represent arbitrarily complex distributions but require large datasets and substantial compute. Many scientific problems occupy an intermediate regime: low- to moderate-dimensional outputs with structured, multimodal uncertainty. For these problems, explicit mixture-based models balance expressivity, statistical efficiency, and interpretability.

% Scientific machine learning (SciML) increasingly operates in regimes where deterministic prediction is neither possible nor meaningful. Many physical systems are noisy, partially observed, chaotic, or under-specified, so that a single input admits multiple valid outcomes. In these settings, the goal of learning is not to approximate a pointwise input--output map, but to characterize the full conditional distribution of outcomes. This requires a shift from point-estimate accuracy toward principled uncertainty quantification (UQ).

% Most deep learning methods remain rooted in point estimation, typically trained by minimizing Mean Squared Error (MSE). While effective for well-posed deterministic problems, this paradigm implicitly treats deviations from the mean as noise to be averaged away. In scientific applications, however, variability often reflects genuine physical ambiguity. Averaging across distinct outcomes can therefore produce predictions that are unphysical, misleading, or incompatible with downstream scientific analysis.

A central distinction in UQ is between epistemic and aleatoric uncertainty. Epistemic uncertainty arises from limited data or model misspecification and can, in principle, be reduced by collecting additional observations. Aleatoric uncertainty is irreducible: it reflects inherent stochasticity, unresolved degrees of freedom, or latent variables absent from the model. In many scientific problems, multimodality is a manifestation of aleatoric uncertainty. Even with infinite data and a perfect model, the conditional distribution \( p(y \lvert x) \) may remain multimodal. Methods that primarily target epistemic uncertainty, such as deep ensembles and BNNs, are generally incapable of capturing aleatoric uncertainty. This work focuses on modeling the structure of the aleatoric solution landscape itself.

\subsection{Limitations of Unimodal Predictive Models}

Most UQ methods in regression implicitly assume unimodal predictive distributions, often Gaussian. This assumption is deeply limiting in several scientific settings.

Gaussian Processes, while theoretically rigorous, impose Gaussian priors and likelihoods, leading to unimodal posterior predictions that smear probability mass across distinct outcomes \cite{RasmussenWilliams2006GaussianProcesses}. Similarly, Bayesian neural networks (BNNs) in practice tend to collapse multimodal distributions over outputs into broad unimodal densities, even if the distributions over weights is multimodal \cite{Izmailov2021WhatAreBNNPosteriors}. Deep ensembles improve robustness and out-of-distribution detection, but when trained with MSE objectives still concentrate around the conditional mean, capturing epistemic uncertainty rather than the structure of the data-generating process \cite{DeepEnsembles2021}.

%These limitations become critical in several scientific settings:
\paragraph{Ill-posed inverse problems.} Inverse mappings are frequently one-to-many. A single observation may correspond to multiple physically valid configurations, as in nonlinear inverse regression problems. Unimodal predictors return convex averages of these solutions, producing estimates that may not correspond to any realizable physical state.

\paragraph{Chaotic dynamical systems.} In systems with positive Lyapunov exponents, such as low-dimensional chaotic flows, long-term pointwise prediction is fundamentally impossible. Meaningful forecasts must represent distributions over trajectories rather than single paths.

\paragraph{Multistable systems.} Many physical systems exhibit multiple stable attractors separated by basins of attraction. Noise or small perturbations can determine which attractor is reached, making the conditional outcome distribution multimodal, even if the underlying dynamics are deterministic.

\section{A Probabilistic Framework for Scientific Learning}
\label{sec:probabilistic-framework}

To unify the discussion of modeling choices in SciML and to clarify where Mixture Density Networks fit relative to existing approaches, we introduce a probabilistic framework that categorizes models by how they represent conditional uncertainty. This framework is intentionally minimal, yet sufficient to explain the differences between approaches.

\subsection{From Point Prediction to Conditional Distributions}

Let $\mathcal{D} = \{(x_i, y_i)\}_{i=1}^N$ be a dataset of input–output pairs, where $x \in \mathcal{X}$ are control parameters, initial conditions, or observations, and $y \in \mathcal{Y}$ denotes a quantity of interest.

The dominant paradigm in deep learning-based regression is point prediction: a deterministic map of the form
\begin{equation}
    f_\theta : \mathcal{X} \to \mathcal{Y},
\end{equation}
trained by minimizing a pointwise loss such as Mean Squared Error. This approach implicitly assumes unimodal uncertainty, well summarized by the conditional mean. While appropriate for well-posed deterministic problems, this assumption is violated whenever uncertainty reflects genuine physical ambiguity rather than measurement noise.

A probabilistic SciML model instead aims to learn the full conditional distribution $p(y \lvert x)$. We refer to any such model as a \emph{distribution network}. Distribution networks differ primarily in how this conditional distribution is represented and how probability mass is allocated during training.

\subsection{Explicit and Implicit Distribution Networks}

Distribution networks can be broadly categorized into two classes: \emph{explicit} and \emph{implicit} models.

\paragraph{Explicit distribution networks.}
Explicit models parameterize the conditional probability density function directly and optimize it via maximum likelihood. Formally, they output a tractable density $p_\theta(y \lvert x)$ and minimize the negative log-likelihood (NLL),
\begin{equation}
    \mathcal{L}(\theta) = -\frac{1}{N} \sum_{i=1}^N \log p_\theta(y_i \lvert x_i).
\end{equation}

The defining characteristic of explicit models is that probability mass allocation is an explicit part of the objective. Errors in mode weights, tail behavior, or low-probability regions are directly penalized. As a result, these models are naturally suited to settings where downstream decisions depend on well-calibrated likelihoods rather than point predictions or samples alone.

\paragraph{Implicit distribution networks.}
Implicit models define the conditional distribution through a sampling procedure,
\begin{equation}
    y = G_\theta(x, \varepsilon), \qquad \varepsilon \sim p(\varepsilon),
\end{equation}
without requiring a closed-form expression for the density. This class includes GANs, conditional variational autoencoders \cite{sohn2015CVAE}, diffusion models \cite{ho2020ddpm}, and Conditional Flow Matching \cite{lipman2022flow}.

Implicit models are highly expressive and can approximate complex distributions in high-dimensional settings. However, because the density is not directly optimized, training relies on surrogate objectives such as adversarial losses or score matching. These objectives prioritize local geometric fidelity of samples, rather than direct control over global probability mass allocation.

\subsection{Mixture Density Networks as Structured Conditional Models}

Within the class of explicit distribution networks, Mixture Density Networks (MDNs) parameterize the conditional density as a finite mixture of Gaussian distributions:
\begin{equation}
    p_\theta(y \lvert x) = \sum_{k=1}^K \alpha_k(x;\theta)\, \mathcal{N}\!\left(y \lvert \mu_k(x;\theta), \Sigma_k(x;\theta)\right),
\end{equation}

where the mixture weights $\alpha_k(x)$ form a simplex (that is, $\sum \alpha_k = 1$ and $\alpha_k \ge 0$) and are learned jointly with the component parameters $\mu_k$ and $\Sigma_k$. The distribution parameters $\{\alpha_k, \mu_k, \Sigma_k\}$ are outputs of a neural network backbone.

This formulation encodes a minimal but powerful structural assumption: uncertainty can be decomposed into a small number of distinct outcomes, each well approximated by a Gaussian. Importantly, the representation separates \emph{where} probability mass lies (component means and covariances) from \emph{how much} mass each carries (mixture weights).

MDNs thus combine the calibration benefits of explicit likelihood optimization with structural priors suited to problems where uncertainty decomposes into discrete, interpretable branches. Their key distinction from modern generative models is inductive bias: MDNs decompose uncertainty into discrete components, while implicit models offer no such guarantee. For many scientific problems with structured uncertainty, moderate dimensionality, and limited data, this yields a strong trade-off between expressivity, efficiency, and interpretability, motivating MDNs as a principled foundation for uncertainty-aware SciML. Appendix \ref{sec:mdn4sciml} provides more insights and a fuller description of the MDN implementation for computational scientists, including a discussion on how to select the number $K$ of mixture components.

\section{Inductive Biases in Multimodal Scientific Learning}
\label{sec:inductive-bias}

Scientific machine learning problems often operate in a regime characterized by low-to-moderate dimensional outputs, structured multimodal uncertainty, and limited data. 
This regime -- outputs in $\mathbb{R}^d$ with $d \le 20$, structured multimodal uncertainty, and training sets of $N < 10{,}000$ -- characterizes a broad class of practical scientific problems, including reduced-order modeling, parameter estimation, and state inference in dynamical systems. In such settings, performance is governed less by raw expressivity and more by inductive bias. In this section, we argue that Mixture Density Networks and modern diffusion or flow-based models embody fundamentally different inductive assumptions, leading to systematically different behavior in data efficiency, topological fidelity, and interpretability.

\subsection{Sample Efficiency of Explicit Versus Implicit Distribution Networks}

The explicit/implicit distinction introduced in Section  \ref{sec:probabilistic-framework}  has direct statistical consequences. MDNs are \emph{parametric} estimators: they assume that the true conditional density lies within, or is well-approximated by, a finite family of mixture distributions. When this assumption is satisfied, classical results from statistical learning theory guarantee parametric convergence rates of order $O(n^{-1/2})$ for maximum likelihood  \cite{van2000asymptotic}. Recent advances refine these guarantees for Gaussian mixtures, showing that the sample complexity to achieve error $\epsilon>0$ scales polynomially with output dimension $d$ and number of components $K$, on the order of $\tilde{O}(K d^2 / \epsilon^2)$ \cite{Ashtiani2018Nearly,Suresh2014Near}. This regime is particularly relevant for SciML, where outputs are often low-dimensional quantities such as positions or reduced-order summaries.

Implicit generative models operate under a fundamentally different regime. Diffusion and flow-based models act as \emph{non-parametric} density estimators, learning either a continuous transport map or a score field over the ambient space. In the absence of strong structural assumptions on the target distribution, non-parametric estimation is known to suffer from slower convergence rates of the form $O(n^{-s/2(d+s)})$, where $s$ reflects smoothness assumptions \cite{Wasserman2006Nonparametric}. Recent research proved theoretically and through experiments that the sample complexity of score-based methods such as diffusion and flow matching is generally $\tilde{O}(n^{-1/2(d+5)})$ \cite{chen2023score}.

This distinction becomes critical in data-limited settings. Since implicit models learn through local updates, they require sufficient samples to populate not only the high-density regions but also the paths that connect them. In contrast, explicit likelihood-based models can reassign probability mass globally through their parameters and have built-in assumptions about the decay rate of distributions. In MDNs, a single observation from a rare mode induces a direct gradient signal on the corresponding mixture weight, immediately correcting the estimated mixing proportions. As a result, MDNs recover multimodal structure with far fewer samples, given the mixture assumption is reasonable.

This contrast highlights that the relative strengths of MDNs and diffusion models arise from inductive bias rather than expressivity. Explicit mixture models encode the assumption that uncertainty is structured and decomposable into a small number of components. For many scientific inverse problems and dynamical systems, this assumption reflects physical structure and leads to superior statistical efficiency.
% Our experiments in section \ref{sec:experiments} demonstrate this statistical advantage for data-scarce settings, with MDNs consistently outperforming implicit flow-based models over a diverse range of problems.

\subsection{Likelihood Artifacts in Implicit Distribution Networks}

A central difficulty for score-based and flow-based models arises when the target distribution has disconnected or weakly connected support. Recent analyses of score matching show that the sample complexity depends inversely on the isoperimetric constant of the target distribution \cite{Koehler2023Statistical}. Intuitively, when modes are separated by low-density regions, the score signal between them is weak or ill-defined. As a result, diffusion and flow models can accurately learn the overall shape of each mode while failing to estimate their \emph{relative weights}, leading to systematic overestimation of low-probability regions.

This phenomenon is closely tied to topology. Flow-based models are constructed as smooth invertible maps and therefore preserve topological invariants such as connectedness \cite{Flouris2023Canonical,Cornish2020Relaxing}. When the base distribution is connected but the target distribution is not, continuous transport cannot separate modes without introducing artificial probability ``bridges'' or highly stiff vector fields. These artifacts correspond to physically implausible states in many scientific applications. Figure \ref{fig:data-efficiency-inverse-sine} shows this in practice by consistently overestimating the likelihood of rare events located in between high-likelihood modes. See Appendices \ref{sec:flow-topology} \& \ref{sec:isoperimetry} for an extended analysis.

MDNs avoid this issue by construction. Because they optimize the NLL of an explicit mixture distribution, probability mass can be reassigned globally through the mixture weights without traversing low-density regions. Additionally, the individual Gaussian components have the strong yet reasonable assumption that probability distributions decay exponentially, helping capture low-likelihood regions.

\subsection{Interpretability in MDNs}

Another important consequence of the explicit, parametric structure of Mixture Density Networks is interpretability. By modeling the conditional distribution as a finite mixture, MDNs decompose uncertainty into a small number of distinct components. In scientific settings, these components, represented by $\mu_k(x)$ and $\sigma_k(x)$, often align with meaningful solution branches, while the corresponding mixture weights $\alpha_k(x)$ encode their relative likelihood as a function of the input $x$. Concretely, the scalar field $\alpha_k(x)$ acts as a learned order parameter: regions where $\alpha_k(x)\approx1$ correspond to deterministic outcomes, while transitions between dominant components identify phase boundaries or separatrices. This structure is directly accessible from the trained model and does not require auxiliary losses, post hoc probing, or Monte Carlo analysis. As a result, MDNs provide a transparent view of how uncertainty is organized, rather than only how samples are generated. In contrast, even when implicit generative models produce high-quality samples, they do not expose the probability structure in a directly usable form, making scientific interpretation dependent on additional sampling and analysis.

An example of this phenomenon can be seen in the gravitational attractor example outlined in section \ref{sec:gravitational-attractor}, where the MDN learns to dedicate a mixture element to each of the three possible attractors. Additionally, the input dependence of the weight of the mixture $\alpha_k(x)$ corresponding to each attractor reveals an interpretable structure across a range of scenarios, as can be seen in Figure \ref{fig:attractor_interpretability}.

\input{Text/experiments_section}

\section{Conclusion}

\paragraph{Summary. }
This work revisits Mixture Density Networks as a principled approach for modeling multimodal conditional uncertainty in scientific machine learning. While implicit generative models such as diffusion and flow matching are highly expressive, they often exhibit high sample complexity and topological biases that produce unphysical artifacts in scientific settings. In contrast, the parametric inductive bias of MDNs enables statistically efficient density estimation in data-limited regimes common in SciML. Their mixture structure yields interpretable components that align naturally with physical regimes and phase transitions.

\paragraph{Limitations. }
Our experimental comparisons focus on CFM as a representative implicit generative model; a comprehensive comparison including normalizing flows, conditional VAEs, and probabilistic ensembles would further clarify the trade-offs but is beyond the scope of this work. 
Selecting an appropriate number of mixture components introduces a modeling choice that may require validation or domain knowledge, and fixed parametric components can be restrictive for very high-dimensional or weakly structured uncertainty. While MDNs are well suited to moderate-dimensional scientific outputs, scaling explicit mixture models to large, unstructured spaces remains an open challenge. These considerations position MDNs not as a universal replacement for implicit generative models, but as a complementary tool whose strengths are closely matched to the structure of many scientific problems.

\paragraph{Future Directions. }
Promising directions include decoupling the MDN probabilistic framework from specific architectures by pairing it with more expressive or domain-aligned backbones. Prior successes with RNN-MDNs in sequence modeling demonstrate that explicit mixture outputs integrate naturally with structured networks while preserving interpretability and calibrated uncertainty \cite{ha2017SketchRNN, ha2018WorldModels}. Within AI for Science, extending this paradigm to operator learning offers a path toward uncertainty-aware neural surrogates for PDEs \cite{thakur2024md-nomad}, where multimodal solution operators arise naturally from bifurcations and parameter ambiguity. Finally, integrating MDNs with Kolmogorov-Arnold Networks \cite{liu2024kan} or variant architectures \cite{guilhoto2025actnet}, may open new opportunities for interpretability in low-dimensional problems.

%% file: Text/experiments_section.tex
\section{Experiments}
\label{sec:experiments}

We empirically validate our claims regarding sample complexity, topological robustness, and interpretability using a diverse suite of experiments. 
We compare Mixture Density Networks (MDNs), Conditional Flow Matching (CFM), and Mean Squared Error (MSE) across four problems exhibiting intrinsic multimodality. We select CFM as representative of implicit generative approaches because it reflects current best practices in transport-based conditional generation and shares the core continuous-transport mechanism with diffusion models.
Full experimental details can be found in Appendix \ref{sec:appendix-experiments}.

\subsection{Sinusoidal Inverse Problem}
\label{sec:sinusoidal}

We begin with an inverse sinusoid benchmark inspired by \cite{bishop1994mixture}. Consider the forward process defined by $y = f(x) + \epsilon$, where $f : \mathbb{R} \rightarrow \mathbb{R}$ is given by $f(x) = \tfrac{x}{2} + 0.7 \sin(5x)$ and $\epsilon \sim N(0, \sigma^2)$ with $\sigma = 0.2$. Given data generated by this process, our goal is to infer the conditional density $p(x \lvert y)$ from noisy observations, rather than to recover a single inverse solution.

Although low-dimensional, this problem captures a defining feature of scientific inverse problems: the forward map is well-posed and unimodal, but its inverse is inherently one-to-many. As a result, deterministic regression trained with MSE collapses the solution set to a single conditional expectation, producing estimates that do not correspond to any realizable solution. This failure mode is illustrated in Figure \ref{fig:mse_inverse_sine} (Appendix), where the model predicts averages across incompatible branches of the inverse mapping.

\begin{figure}[ht]
    \centering
    \includegraphics[width=0.99\linewidth]{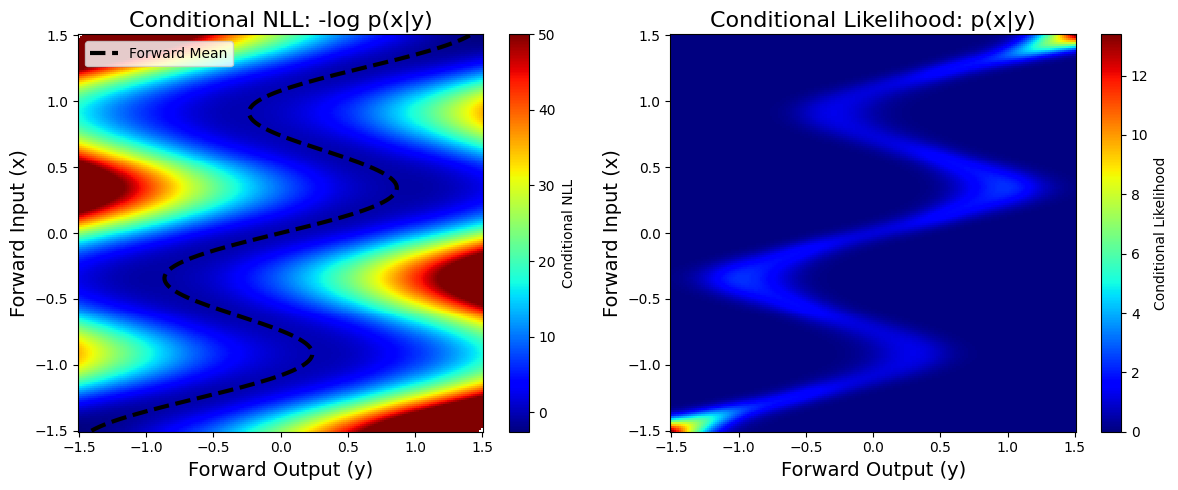}
    \caption{True conditional density of $x$ given $y$ in the inverse sine problem. (left) Conditional NLL. (right) Conditional likelihood.}
    \label{fig:inverse_sine_gt_distribution}
\end{figure}

\begin{figure}[ht]
    \centering
    \includegraphics[width=0.99\linewidth]{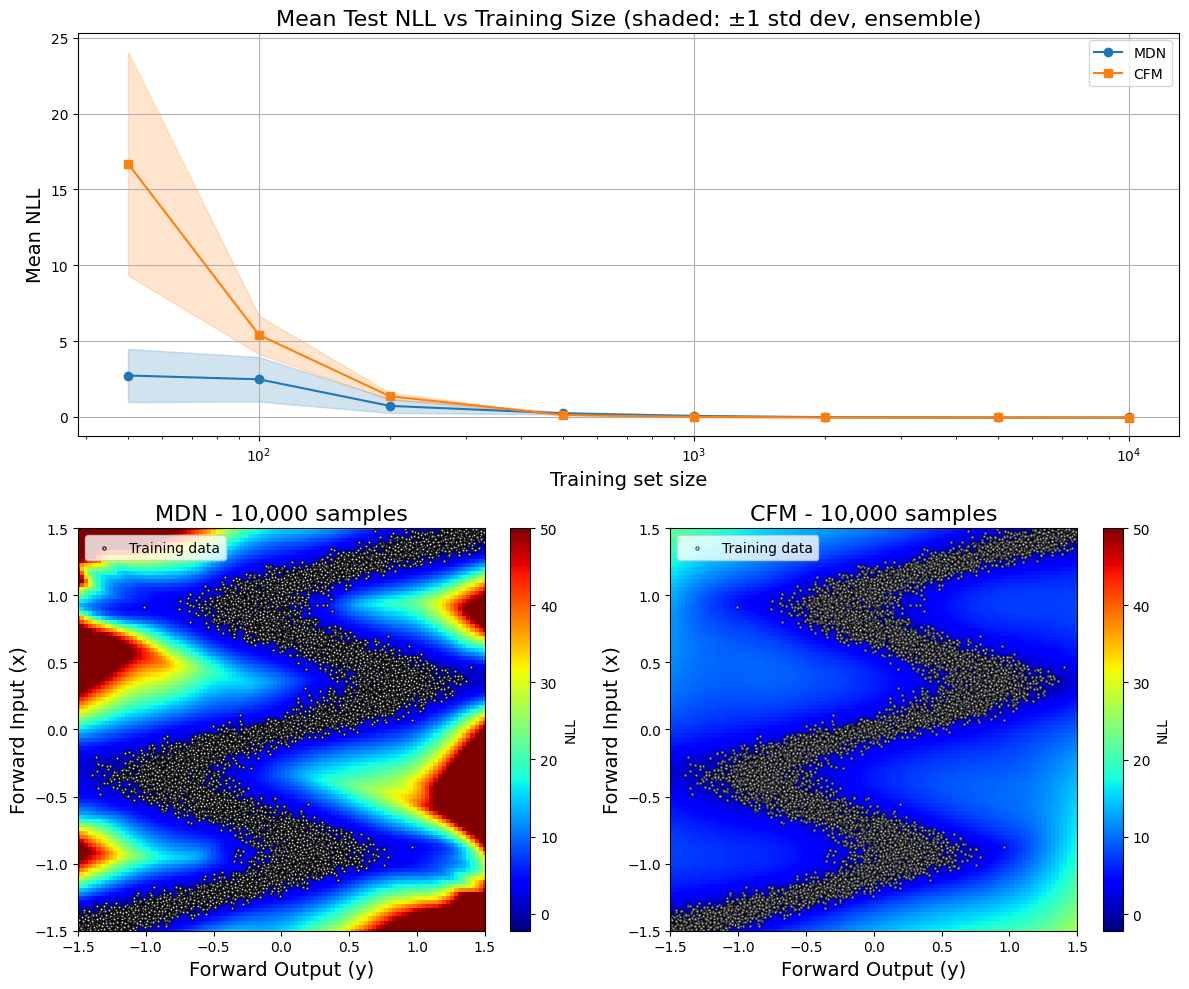}
    \caption{Comparison of MDN and CFM on the sinusoidal inverse problem. Top: test NLL versus training set size, with bands showing 1 Std. Dev. over 12 seeds. Bottom: averaged NLL surfaces.}
    \label{fig:data-efficiency-inverse-sine}
\end{figure}

The geometry of the inverse problem is shown in Figure \ref{fig:inverse_sine_gt_distribution}, where the solution manifold traces a folded, non-injective curve. In the central region of the domain, a single observation $y$ admits multiple disjoint pre-images under $f$, yielding a  multimodal conditional density. Because this conditional structure is well approximated by a finite mixture of smooth components, MDNs provide a natural parametric inductive bias. Consistent with this perspective, MDNs achieve substantially lower error in the low-data regime (shown in Figure \ref{fig:data-efficiency-inverse-sine}), with estimation accuracy improving rapidly as the number of samples increases.

In contrast, Conditional Flow Matching, which learns densities through continuous transport without explicit structure, requires significantly more data to achieve comparable fit, as shown in the top panel of Figure \ref{fig:data-efficiency-inverse-sine}. While both approaches converge as the training set grows to $N = 10{,}000$, CFM exhibits a persistent tendency to introduce over-smoothed probability bridges between disconnected solution branches. This artifact is not specific to CFM; any continuous normalizing flow faces similar topological constraints. This effect is visible in the conditional NLL in Figure \ref{fig:data-efficiency-inverse-sine}, where CFM spreads mass across low-probability regions, overestimating rare events and artificially reducing the negative log-likelihood. These results empirically support the parametric efficiency and topological alignment of MDNs argued in Section \ref{sec:inductive-bias}, highlighting their advantage for inverse problems with disconnected solution sets.

\subsection{Partially Observed Gravitational Attractor}
\label{sec:gravitational-attractor}

In Section \ref{sec:inductive-bias}, we argued the importance of interpretability in scientific machine learning as the ability to recover meaningful macroscopic structure from a model’s internal representation. In this view, the mixture weights $\alpha_k(x)$ of an MDN act as learned order parameters, encoding phase structure and basin membership directly from data. We validate this claim on a simple dissipative dynamical system with multiple stable attractors.

We simulate the motion of a particle in a two-dimensional plane under gravity from up to three fixed celestial bodies located on the unit circle. The system includes a drag term, making it dissipative: all trajectories eventually converge to one of the attractors. Initial positions are sampled uniformly from a disk of radius $0.2$, and small isotropic observation noise is added. The learning task is to predict the noisy discretized trajectory given the initial position.

To isolate the effect of different sources of uncertainty, we train a separate MDN for each of three experimental settings. Across all cases, each predicted mixture mean $\mu_k(x)$ spontaneously learns to encode trajectories towards an individual attractor. In doing so, the model distinguishes separate, discrete possibilities for the final position of the particle.

In the first scenario we considered, each trajectory evolves under the influence of a single attractor chosen at random. The MDN correctly learns input-independent mixture weights, reflecting that the final outcome is decoupled from the initial position $x$. In the second scenario, all attractors are active, but the initial position is perturbed by noise. Here, the learned mixture weights reveal regions of high uncertainty near the origin, where basins overlap, and increasingly sharp assignments farther away, where outcomes become more predictable. In the third setting, all attractors are active and there is no input noise. Even in this deterministic case, the MDN partitions the state space into distinct basins of attraction, effectively behaving as a mixture-of-experts model \cite{jordan1994moe-and-em, zhou2022moe} that learns the underlying dynamics of the system.

\begin{figure*}[ht]
    \centering
    \includegraphics[width=0.32\textwidth]{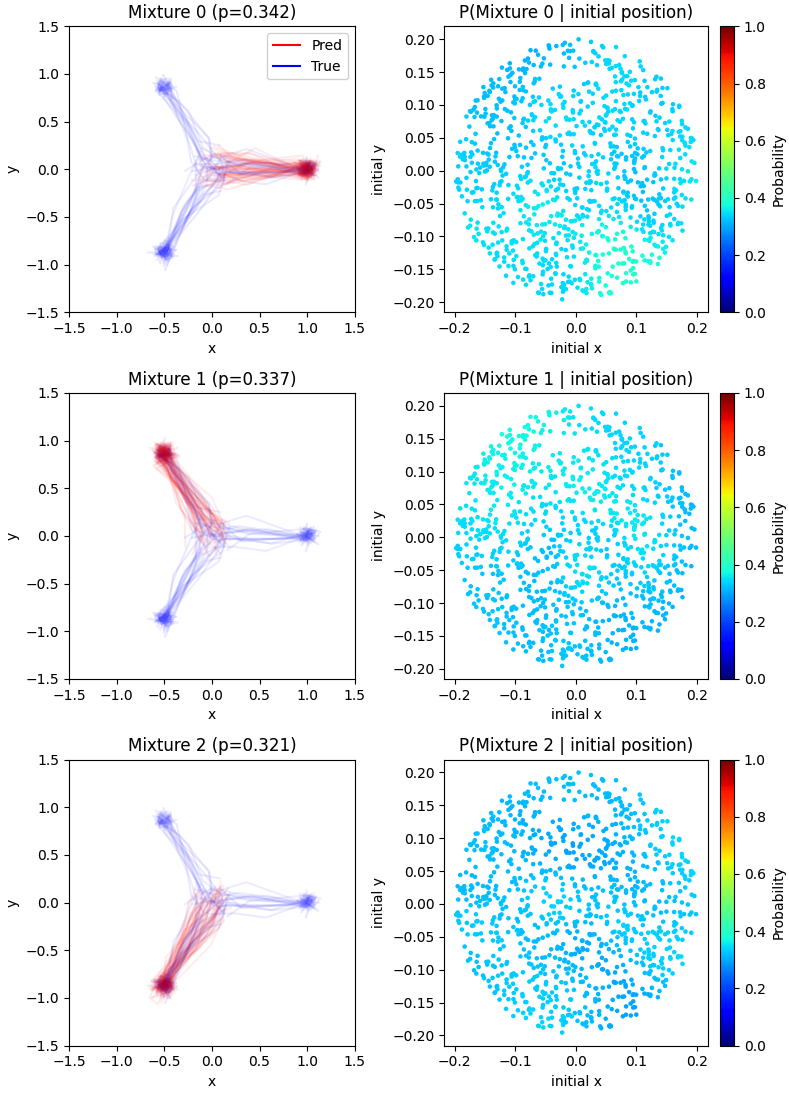}
    \vline
    \includegraphics[width=0.32\textwidth]{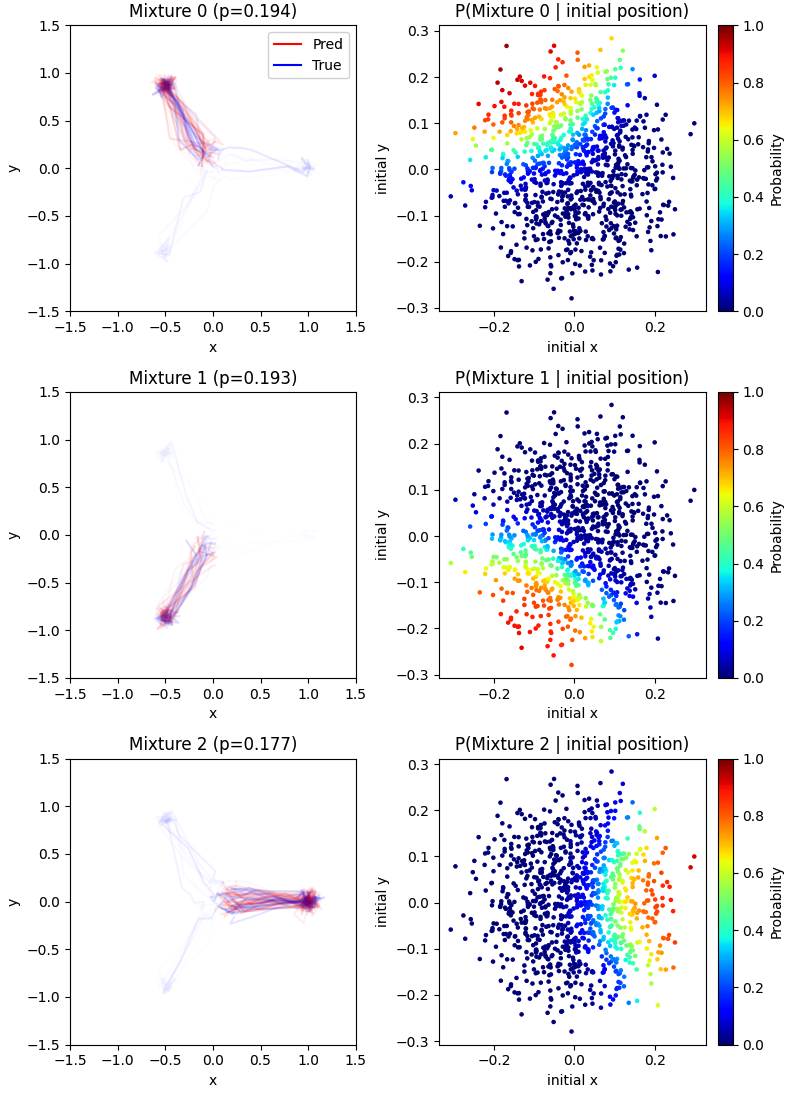}
    \vline
    \includegraphics[width=0.32\textwidth]{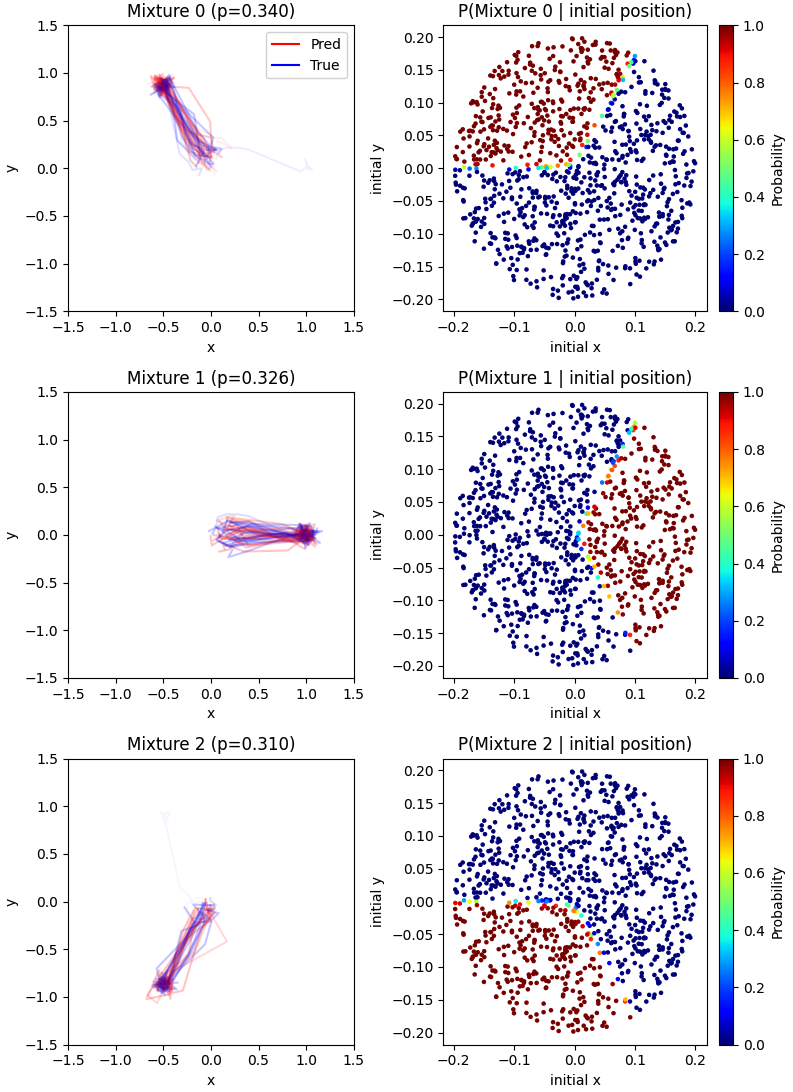}
    \caption{Top three mixture elements (by marginal probability) for the three scenarios in the gravitational attractor example. The left columns show mixture element samples, while the right show the weight of that mixture element given the input. Sample opacities are weighted by mixture probability. (left) Case 1. (middle) Case 2. (right) Case 3.}
    \label{fig:attractor_interpretability}
\end{figure*}

Figure \ref{fig:attractor_interpretability} visualizes the dominant mixture components and their associated weights across all three scenarios. In contrast, a Conditional Flow Matching model trained on the same data offers no mechanism to recover basin structure or decision boundaries. Across all three problems, MDNs outperform CFM in test NLL (shown in Table \ref{tab:gravitational_attractor_test_nll}). This comparison reinforces a central claim of this work: the parametric mixture structure of MDNs provides efficiency and semantic interpretability, yielding insights that implicit generative models do not readily expose.

\subsection{ODE Saddle-Node Bifurcation}

Bifurcations in dynamical systems represent critical transitions where small coefficient changes drastically alter behavior. In under-specified systems where these coefficients are hidden, accurate uncertainty quantification is required to determine possible trajectories \cite{hendriks2025equivariantflowmatchingsymmetrybreaking, psarellis2025active}.

The saddle-node bifurcation is one of the most fundamental bifurcations in dynamical systems. We consider the canonical form (further described in Appendix \ref{sec:appendix-experiments}), where $x$ is the state variable and $r \in \mathbb{R}$ is the bifurcation parameter,
\begin{equation}
\frac{dx}{dt} = r + x^2.
\label{eq:saddle_node}
\end{equation}
The system's behavior strongly depends on $r$. For $r<0$, two fixed points exist at $x^* = \pm\sqrt{-r}$, with the point $-\sqrt{-r}$ being stable (attracting), and $+\sqrt{-r}$ being unstable (repelling). For $r = 0$, a semi-stable fixed point exists at $x^* = 0$ (the saddle-node bifurcation point). For $r > 0$, no fixed points exist, and all trajectories diverge to infinity.

For this problem, we treat the bifurcation parameter $r$ as a latent variable: the model only receives the initial condition $x_0$ as input. This creates an under-specified problem where the model must learn to predict a distinct trajectory for each possible $r$, which yields a strongly multimodal distribution.

\begin{figure}[ht]
    \centering
    \includegraphics[width=0.9\linewidth]{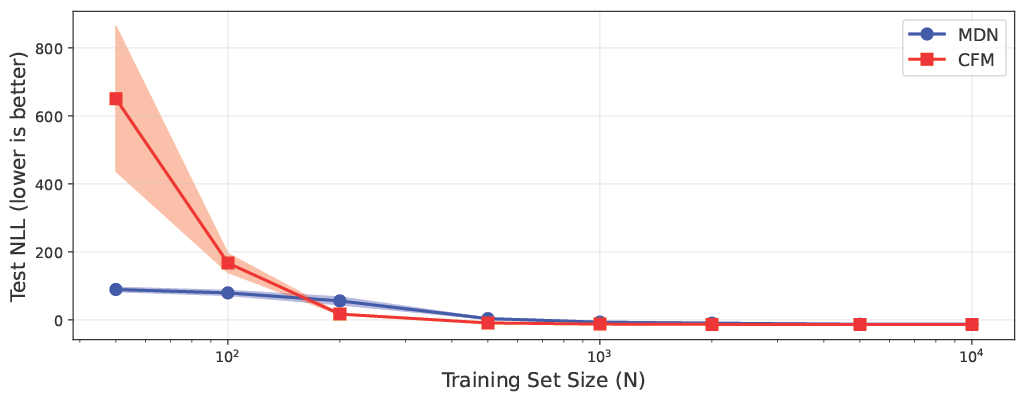}
    \caption{Data efficiency comparison in the bifurcation system. Shaded regions indicate $\pm 1$ Std. Dev. of 12 ensemble members.}
    \label{fig:saddle_node_efficiency}
\end{figure}

Figure~\ref{fig:saddle_node_efficiency} presents the sample efficiency comparison between MDN and CFM. Both methods successfully capture the multimodal nature of this problem, correctly identifying different trajectories. However, MDN shows a clear advantage in the low-data regime: at $N = 50$ training samples, MDN achieves a test NLL of $85.6 \pm 6.5$, compared to $691.2 \pm 146.5$ for CFM. 
This order-of-magnitude advantage persists across hyperparameter configurations and random seeds, confirming it reflects sample complexity rather than optimization failure. The gap is practically significant: in many experimental settings, collecting even 50 labeled trajectories may be costly.
With more data ($N = 10{,}000$), both methods achieve similar performance.

A standard MSE-trained network fails on this task by predicting the conditional mean, an average of the divergent and convergent trajectories, while the true trajectories exhibit clear multimodal behavior. This ``averaging''  behavior is a fundamental limitation of point-estimate models: when multiple disconnected outcomes are possible, the mean prediction may lie in a region of low probability density, producing physically implausible forecasts. In contrast, MDN and CFM correctly capture both modes, generating samples that track both the divergent and convergent trajectories.

\subsection{Lorenz System}\label{sec:lorenz}

Finally, we evaluate the ability of different uncertainty-aware models to capture the chaotic dynamics of the Lorenz–63 system \cite{lorenz1963deterministic}. In this regime, infinitesimal perturbations to the initial position grow exponentially over time, rendering pointwise long-horizon prediction fundamentally ill-posed. This behavior is exacerbated by adding observational noise, effectively guaranteeing diverging predictions even in short time horizons. Meaningful prediction therefore requires modeling the distribution over future trajectories rather than a single deterministic path.

We train autoregressive models to predict state increments $\Delta x_t := x_{t+\Delta t} - x_t$ and assess their long-term behavior via rollouts. We compare a standard neural network trained with MSE, a feedforward Mixture Density Network (MDN), and a recurrent MDN (RNN-MDN) \cite{ha2018WorldModels, ha2017SketchRNN}. As expected, observational noise causes MSE-trained models to fail to reproduce the qualitative structure of the attractor, generating trajectories that drift outward due to averaging over incompatible local dynamics. The feedforward MDN captures local variability but accumulates noise over long horizons. In contrast, the RNN-MDN consistently produces trajectories that preserve the system's true geometric and topological features.

\begin{figure}[ht]
    \centering
    \includegraphics[width=0.5\textwidth]{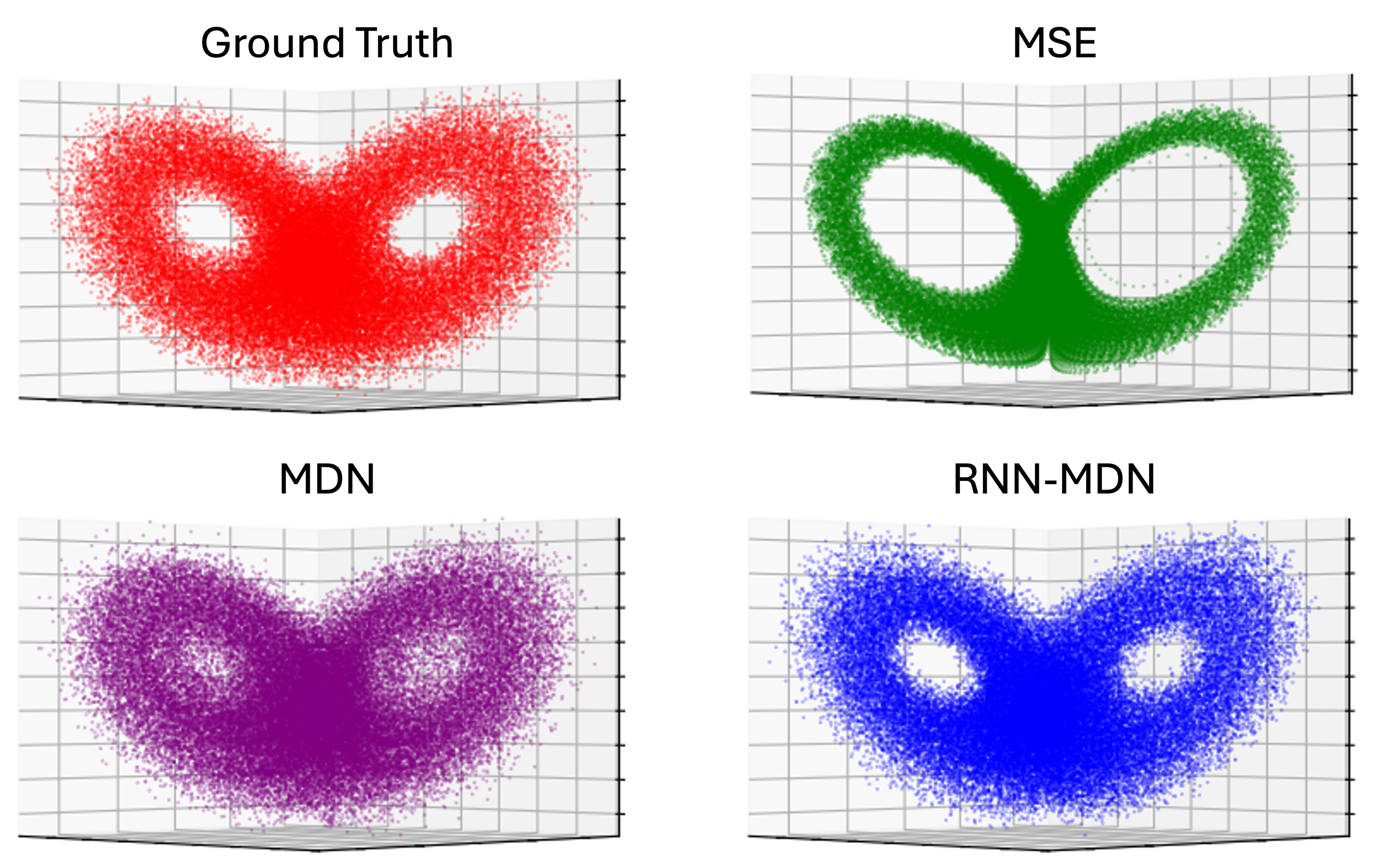}
    \caption{Rollout trajectories in the Lorenz system. Only RNN-MDN recovers the geometry and topology of the true attractor.}
    \label{fig:lorenz-trajectories}
\end{figure}

These differences are evident in Figure \ref{fig:lorenz-trajectories}, where only RNN-MDN recovers the characteristic double-lobed structure of the attractor without artificial distortion. Quantitatively, this behavior is reflected in the Maximum Mean Discrepancy (MMD) between predicted and true trajectories in Table \ref{tab:lorenz-mmd}, where RNN-MDN achieves the lowest discrepancy. 
Notably, RNN-MDN uses fewer parameters (58K) than the MSE baseline (67K). Moreover, the feedforward MDN already achieves a 27x improvement in MMD over MSE despite matching architectures, isolating the contribution of the mixture output head. The additional gains from RNN-MDN demonstrate that explicit multimodal outputs integrate naturally with recurrent structure. RNN-MDN is also capable of partitioning the space into interpretable components, which can be seen in Figure \ref{fig:rnn-mdn-mixture-decomposition} (Appendix).

\begin{table}[ht]
    \centering
    \begin{tabular}{cccc}\hline
        & MSE & MDN & RNN-MDN \\\hline
        MMD ($\downarrow$) & 2.251e-2 & 8.104e-4 & \textbf{3.813e-4}\\ \hline
    \end{tabular}
    \caption{Maximum Mean Discrepancy (MMD) between point clouds of ground truth data and autoregressive predictions from different methods for the Lorenz system. Lower is better.% We consider the family of RBF kernels with scales $\sigma\in [0.1, 50]$, with full results shown in Figure \ref{fig:lorenz-mmd-vs-sigma} (Appendix).
    }
    \label{tab:lorenz-mmd}
\end{table}

A natural implication of these results is that Mixture Density Networks can be effectively paired with problem-specific backbone architectures, allowing domain-informed inductive biases in the dynamics model to be combined with explicit multimodal uncertainty representations without sacrificing stability or interpretability.

%% file: Text/appendix.tex
\clearpage

\appendix

\section{Mathematical Notation}

Table \ref{tab:math-notation} summarizes the symbols and notation used in this work.

\begin{table}[ht]
\caption{Summary of the symbols and notation used in this paper.}
\label{tab:math-notation}
\begin{center}
    \begin{small}
        \begin{sc}
        
\begin{tabular}{ll}\toprule
    \textbf{Symbol} & \textbf{Meaning} \\ \toprule
    $\vx$ & A vector $\vx=(x_1,\dots,x_d)$ in $\R^d$\\\midrule
    $d$ & \makecell[l]{Usually used to denote the\\ dimension of a vector}\\\midrule
    $\mathcal{D}$ & \makecell[l]{A dataset $\mathcal{D}=\{(x_i, y_i)\}_{i=1}^N$ of\\ input/output pairs} \\\midrule
    $N$ & \makecell[l]{Usually used to denote the size\\ of a dataset}\\\midrule
    $\theta$ & Parameters of a neural network\\ \midrule
    $p_\theta$ & \makecell[l]{A probability distribution with\\ learnable parameters $\theta$}\\ \midrule
    $N(\mu, \Sigma)$ & \makecell[l]{A normal distribution with mean\\ $\mu$ and covariance $\Sigma$}\\\midrule
    $\mathcal{N}(y; \mu, \Sigma)$ & \makecell[l]{The PDF of normal distribution \\with mean $\mu$ and covariance $\Sigma$\\ evaluated at a point $y$}\\\midrule
    $K$ & \makecell[l]{The number of mixture elements\\ in a MDN}\\\midrule
    $\alpha_k$ & \makecell[l]{The weight of the $k$th mixture\\ element of an MDN}\\\midrule
    $\mathcal{L}$ & \makecell[l]{A loss function used to train a\\ neural network model}\\\midrule
    $\mI_d$ & The identity matrix of size $d$\\\midrule
    $\mathbb{E}$ & Expectation operand\\ \midrule
    $\texttt{LogSumExp}_k$ & \makecell[l]{The ``log-sum-exponent'' operation\\ over an index $k$}\\ \bottomrule
\end{tabular}
            
        \end{sc}
    \end{small}
\end{center}
\end{table}

\section{The Mixture Density Network for SciML}\label{sec:mdn4sciml}

MDNs have been used across several domains in deep learning in recent years, including image segmentation and object detection tasks \cite{he2020deep, varamesh2020mixture, choi2022md3d}, and for time-series data such as generating realistic-looking handwritten text \cite{ha2017SketchRNN}. In \cite{li2022automotive}, MDNs are used to model radar data from cars to facilitate autonomous driving; in \cite{unni2020deep}, MDNs are used for inverse design of photonic structures; and in \cite{davis2020use} MDNs are used for epidemiological modeling. Despite these use-cases, MDNs have seen little use when it comes to modeling physical systems in the natural sciences.

Having established the theoretical advantages of MDNs for low-dimensional, multimodal scientific tasks, we describe the specific implementation details relevant to computational scientists. 

\subsection{Architecture}

We adopt a standard MDN architecture where the backbone can be any neural network appropriate for the input data type (e.g., MLP for vector inputs, CNN for images/fields \cite{he2020deep, varamesh2020mixture, choi2022md3d, Tosi_2021_CVPR}, RNN/LSTM for time series \cite{ha2017SketchRNN, ha2018WorldModels}, or NOMAD for operator learning \cite{thakur2024md-nomad, seidman2022nomad}). The crucial modification is the output head: for a target $y \in \mathbb{R}^d$ and a mixture of $K$ Gaussians, the network outputs $K(1 + d + d)$ parameters (assuming diagonal covariance) or $K(1 + d + d(d+1)/2)$ (full covariance). Although predicting the full covariance matrix yields more expressive power, in practice most MDN implementations (including the one used in this paper) assumes diagonal covariance.

The network outputs three vectors: logits $z_\alpha\in\R^K$, means $z_\mu\in\R^d$ and scales $z_\sigma\in\R^d$, which are then (possibly) transformed to obtain weights, means and variances, respectively, of the quantity of interest.
\begin{align}
    \text{Logits } z_\alpha & \mapsto \alpha_k = \frac{\exp(z_{\alpha,k})}{\sum_j \exp(z_{\alpha,j})} \quad (\texttt{Softmax}), \\
    \text{Means } z_\mu & \mapsto \mu_k = z_{\mu,k} \quad (\text{Identity}), \\
    \text{Scales } z_\sigma & \mapsto \sigma_k = \log\left(1+\exp(z_{\sigma, k})\right)+ \epsilon \quad (\texttt{Softplus}).
\end{align}
Using a softplus activation plus a small epsilon $\epsilon>0$ for the scale parameters $\sigma$ is preferred over the exponential function to strictly enforce positivity while avoiding numerical explosion (gradients of $e^x$ can be unstable). Another option is to use $z_\sigma \mapsto 1+ELU(z_\sigma) + \epsilon$, where $ELU$ is the Exponential Linear Unit \cite{clevert2016elu}.

\subsection{Numerically Stable Computation of the MDN Loss}

Given a target $y\in\mathbb{R}^d$, the conditional likelihood modeled by the MDN is
\begin{equation}
p(y \mid x) = \sum_{k=1}^K \alpha_k(x)\,
\mathcal{N}\!\left(y \mid \mu_k(x), \sigma_k^2(x)\right),
\end{equation}
and training proceeds by minimizing the negative log-likelihood (NLL).

Direct evaluation of this expression is numerically unstable due to exponentials and potentially small variances. Instead, we compute the NLL using the \texttt{LogSumExp} trick. Defining
\begin{align*}
F_k(x,y) &= z_{\alpha,k}(x) - \texttt{LogSumExp}_j\!\left(z_{\alpha,j}(x)\right) -\log\left(\mathcal{N}(y|\mu_k(x), \Sigma_k(x)\right)
\end{align*}
the loss for a single observation $(x,y)$ can be written compactly as
\begin{equation}
\mathcal{L}(x,y)
= - \texttt{LogSumExp}_k \left( F_k(x,y) \right).
\end{equation}

This formulation, adapted from \cite{Ha-mdn-jax-tutorial}, avoids explicit normalization of mixture weights and prevents numerical underflow when evaluating low-probability components. In practice, a \texttt{LogSumExp}-based implementation is essential for stable MDN training, particularly when mixture weights are highly imbalanced or component variances are small.

The \texttt{LogSumExp} operation is stably computed as 
\begin{align*}
    \texttt{LogSumExp}_k \left( a_k \right) &:= \max_k(a_k) + \log\left(\sum_k \exp\left(a_k- \max(a_k)\right) \right).
\end{align*}

\subsection{Interpreting the Latent Structure}

A distinct advantage of MDNs is the interpretability of the mixture weights $\alpha_k(x)$. In scientific contexts, these weights effectively learn something analogous to a phase diagram of the system.

The scalar field $\alpha_k(x)$ acts as an order parameter. Regions where $\alpha_k(x) \approx 1$ correspond to stable phases. The transition regions where weights shift from one component to another (e.g., $0 \ll \alpha_1 \approx \alpha_2 \ll 1$) identify the separatrices or phase transitions.
By computing the entropy of the mixing weights, $H(\alpha(x))$, one can automatically detect regions of phase coexistence.

This direct access to the internal ``belief state'' of the model is absent in implicit generative models, where the probability density is only accessible via expensive Monte Carlo sampling or ODE trace estimation.

\subsection{Inference Latency and Real-Time Control}

Finally, in applications such as real-time control, inference speed is critical. MDN inference entails $O(1)$ complexity, since a single forward pass of the neural network yields the analytic parameters $(\alpha, \mu, \Sigma)$ of the conditional distribution. On the other hand, DDPM and CFM entails $O(T)$ forward passes, requiring solving an ODE/SDE, typically involving $T=20$ to $T=100$ sequential evaluations of the neural network. Even with recent ``distilled'' solvers \cite{Guo2023Gaussian}, diffusion models remain significantly slower than MDNs. For example, in a low-latency control loop operating at 1.5 kHz \cite{2020adaptiveopticalsystem}, an MDN is often the only viable multimodal probabilistic deep learning option.

\subsection{Ablation on Number of Mixture Elements}
MDNs introduce a new hyperparameter that must be set after choosing a suitable backbone architecture: the number $K$ of mixture components. Experimentally, we have found that there is often little downside to being overcautious and adding more mixture elements than are strictly necessary, as the added computational cost is negligible compared to the backbone network, and the MDN usually learns to not activate unnecessary mixtures. Since in practice it can be hard to know apriori how many components are necessary to describe a system, it can be advantageous to side on the air of caution.

\begin{figure}[ht]
    \centering
    \includegraphics[width=0.99\linewidth]{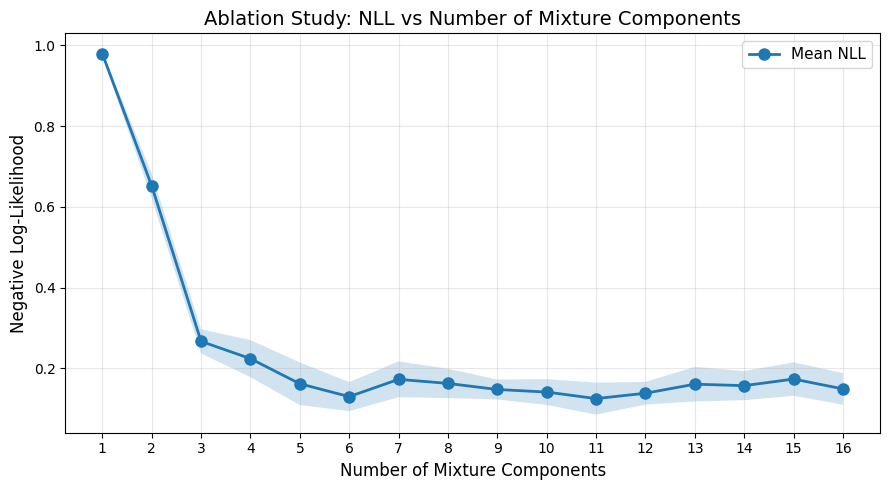}
    \caption{Test NLL for different number of MDN mixture elements in the inverse sinusoid problem. $\pm1$ standard deviation bands are computed across 12 different network initializations.}
    \label{fig:ablation_mixture_elements}
\end{figure}

In Figure \ref{fig:ablation_mixture_elements}, we plot the test NLL for the inverse sinusoid problem described in the main text for MDNs with $K=1,2,\dots, 16$ mixture elements using a dataset of size $N=1{,}000$. As can be seen, after there are enough components to describe the uncertainty in the system (in this case, around 5), there are diminishing returns to increasing $K$, but no performance deterioration.

\begin{figure}[ht]
    \centering
    \includegraphics[width=0.99\linewidth]{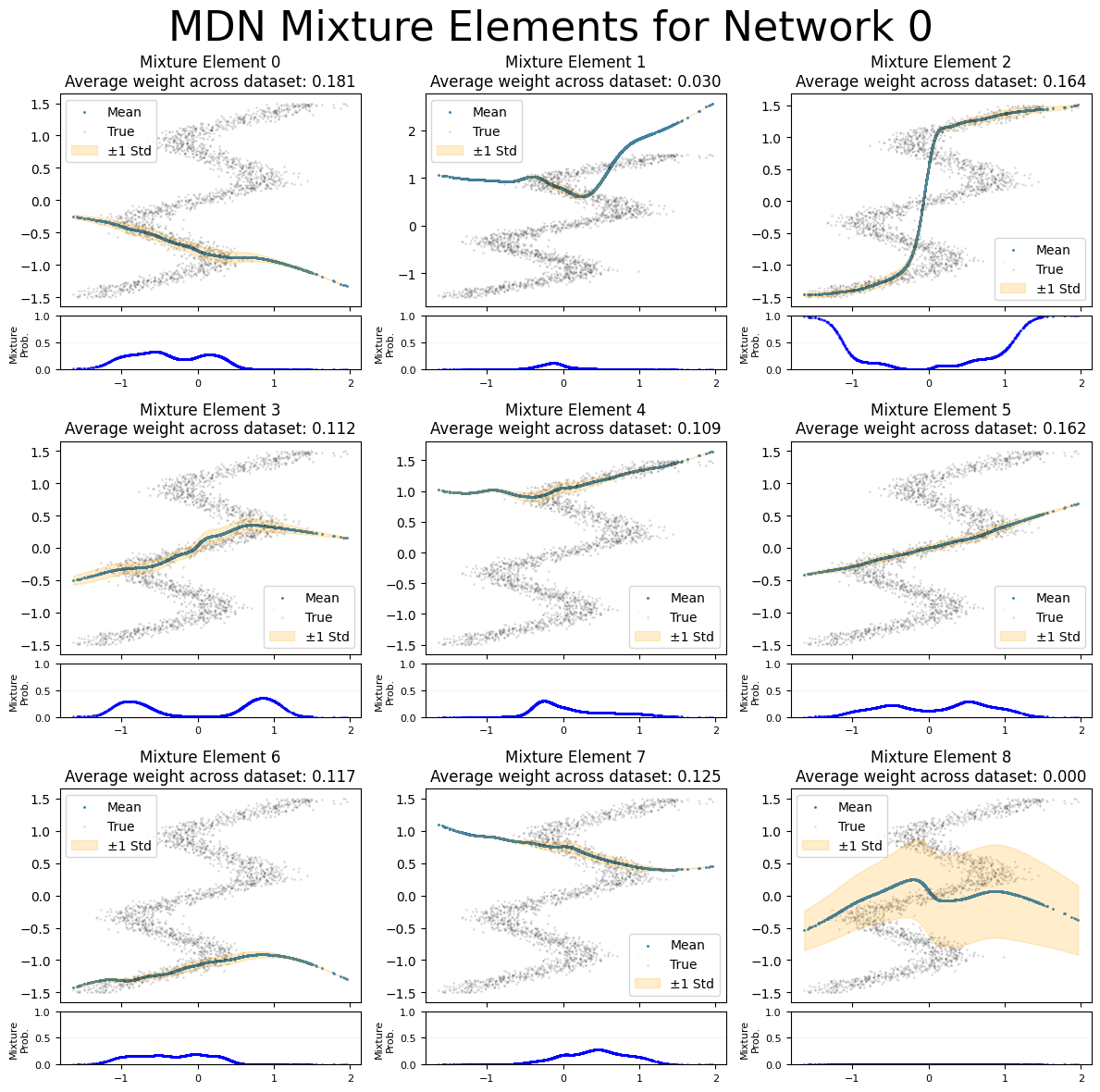}
    \caption{Example of learned mixture components for the inverse sine benchmark with $K=9$ mixture elements. Each subplot displays mean and predicted standard deviation bands for each mixture, along with a smaller plot below showing conditional mixture probabilities given an input ($x$-axis).}
    \label{fig:mixture_elements_inverse_sine}
\end{figure}

The one potential downside to using too many components is that it may be possible for more than one mixture represent the same mode, potentially creating ambiguity and affecting interpretability. For example, in Figure \ref{fig:mixture_elements_inverse_sine}, we can see that mixture element 2 overlaps with mixture elements 4 and 6. In settings where we wish to neatly decompose the domain into ``zones of influence'', this can potentially become a challenge. More often than not, however, unnecessary elements simply get assigned zero or very small weight, as can be seen in mixture number 8 of Figure \ref{fig:mixture_elements_inverse_sine}. Additionally, when overparameterizing the mixture, one can further suppress artifacts at inference time by discarding components whose weights fall below a small threshold (\textit{e.g.} $0.01$), preventing spurious samples from extremely low-probability mixtures\footnote{This feature is implemented in our \texttt{JaxMix} package via the \texttt{restrict\_rare\_event\_rate} option.}. A full description of this ablation, including specific mixture elements can be seen in the \texttt{mixture\_component\_ablation.ipynb} file of our supplementary material.

\section{Aleatoric and Epistemic Uncertainty in Scientific Machine Learning}

Uncertainty in scientific machine learning is commonly decomposed into \emph{aleatoric} and \emph{epistemic} components. Aleatoric uncertainty arises from intrinsic variability in the system or measurement process and persists even with infinite data. Typical sources include sensor noise, stochastic forcing, or genuinely multivalued physical responses, such as those discussed in the main body of this paper. In contrast, epistemic uncertainty reflects incomplete knowledge of the underlying mapping due to finite data and model misspecification and can, in principle, be reduced through additional observations or improved modeling. The contributions in this paper focus specifically on modeling aleatoric uncertainty.

While MDNs primarily target aleatoric uncertainty, epistemic UQ can be addressed through complementary techniques such as ensembles, Bayesian neural networks, or light-weight approaches such as EpiNets \cite{epistemicNNs}. In practice, separating these two sources of uncertainty is essential for scientific decision-making, as they carry different implications for experiment design, model refinement, and physical interpretation.

Epistemic uncertainty also plays a central role in adaptive data acquisition strategies such as Bayesian optimization \cite{wang2022boreview} and active learning \cite{settles2009active}, where the goal is to efficiently reduce model uncertainty through targeted sampling. In SciML, epistemic uncertainty highlights regions of the input space that are underexplored by existing data, guiding where new simulations or experiments are most informative. By explicitly separating aleatoric and epistemic uncertainties, models can avoid oversampling intrinsically stochastic regimes and may instead focus acquisition on regions where additional data meaningfully improves prediction accuracy. This distinction is especially important in high-cost scientific settings, where principled exploration–exploitation trade-offs are critical for efficient discovery, and several papers have targeted contributions to this field in recent years \cite{guilhoto2024neon, Pickering2022, kim2022deep}.

\section{Topological Constraints: Diffeomorphisms vs. Ruptures}\label{sec:flow-topology}

Flow Matching and ODE-based diffusion models are constructed as \textbf{diffeomorphisms}—maps that are differentiable and invertible. A fundamental theorem of differential topology states that diffeomorphisms preserve topological invariants, including the number of connected components \cite{Flouris2023Canonical, Cornish2020Relaxing}.
\begin{equation}
    \text{Topology}(\mathcal{Z}) \cong \text{Topology}(\mathcal{Y}), \quad \text{under diffeomorphism } \phi.
\end{equation}
This creates a profound problem when the base distribution $p(z)$ is a single connected Gaussian and the target physical distribution $p(y|x)$ consists of two or more disconnected peaks. A continuous flow cannot tear the single Gaussian blob into two separate blobs without introducing infinite gradients or singularities in the vector field. To compensate, flow models invariably create thin ``filaments'' or ``bridges'' of non-zero probability connecting the modes. These bridges could represent physically forbidden states (e.g., predicting a particle exists inside a potential barrier). Approximating this ``tearing'' requires the learned vector field to have an extremely high Lipschitz constant, leading to stiff ODEs that are slow and unstable to integrate.  This phenomenon can be seen in Figure \ref{fig:data-efficiency-inverse-sine} in the experiments section, where CFM struggles to learn disconnected modes in the probability landscape.

MDNs, on the other hand, are topologically agnostic in the output space. The mixture distribution is constructed by summing discrete kernels. As the control parameter $x$ crosses a decision boundary, the gating network (which predicts component logits) can quickly switch the active components from one to two smoothly ($K=1 \to K=2$) by modulating the weights $\alpha$. This quasi-discrete ``switching'' capability aligns perfectly with the discrete nature of the problems described in this paper.

\section{The Probabilistic Isoperimetry Problem in Score Matching}\label{sec:isoperimetry}

A critical, often overlooked theoretical limitation of diffusion models in multimodal settings is the \textbf{Probabilistic Isoperimetry Problem}. We use this term to refer to the difficulty of learning distributions with disconnected modes. This has been extensively studied in \cite{Koehler2023Statistical}, which provides a rigorous analysis of the statistical efficiency of score matching. Their work proved that the sample complexity for score matching depends inversely on the \textit{isoperimetric constant} of the target distribution.

Consider a distribution with two modes, $A$ and $B$, separated by a region of negligible probability density (a ``deep energy well''). If  the score function $\nabla \log p(y)$ is the signal used for training, in the empty region between $A$ and $B$, the density is near zero, and the score is flat or ill-defined. Therefore, to learn the relative mass of $A$ vs. $B$ (e.g., $A$ is 90\% likely, $B$ is 10\%), the diffusion process must mix samples between these modes during the noise-injection phase. If the modes are far apart or the dataset is small, the diffusion trajectories rarely cross the low-density barrier efficiently enough to convey the global mixing proportion to the score network. Consequently, score-based models can learn the \textit{shape} of the modes relatively well but fail exponentially in estimating their \textit{relative weights}. This also applies to the low-probability regions between peaks, with methods such as DDPM and CFM overestimating the likelihood of those areas, as further discussed in the previous section.

In contrast, MDNs minimize the NLL directly: $\mathcal{L} = -\sum \log (\sum \alpha_k \mathcal{N})$. This objective function is global. If a data point appears in mode $B$, the loss function heavily penalizes the model if $\alpha_B$ is small, forcing an immediate and direct update to the mixing coefficients. MDNs more effectively ``teleport'' probability mass via the weights $\alpha_k$, bypassing the need to traverse the low-density regions.

\clearpage
\section{Experimental Details}\label{sec:appendix-experiments}

\subsection{Software}
Our code is implemented in JAX \cite{jax2018github} using the Flax \cite{flax2020github} and Optax \cite{deepmind2020jax-optax} libraries to define and train our neural networks. We plan to release our full code base on GitHub if the paper is accepted. In the meantime, we provide it as a ZIP file in the supplementary materials for this submission.

\subsection{Hardware}

Each experiment carried out in this paper was performed using a single NVIDIA A6000 GPU.

\subsection{Computing Negative Log-Likelihood in Flow Matching}

To recover the conditional NLLs for CFM, we were forced to reverse-integrate the flow ODE, along with the divergence of the flow along this path, as described in Appendix C of the original CFM paper \cite{lipman2022flow}. This process is computationally expensive, potentially requiring hundreds of neural network evaluations. In contrast, the MDN yields the full probability distribution in a single forward pass.

To obtain the conditional negative log-likelihood of a trained CFM model, we calculate the log-probability of a data sample $x$ by solving the probability flow ODE backward in time from the data distribution at $t=1$ to the prior noise distribution at $t=0$. This process utilizes the instantaneous change of variables formula, which requires integrating the divergence of the vector field along the sample's trajectory. The final log-likelihood is determined by combining the log-probability of the resulting noise sample $x_0$ under the prior $p_0$, the estimated accumulated divergence, and a log-determinant term that accounts for any data scaling or pre-processing transformations.

Unlike in the original paper that approximates the divergence using the Hutchinson trace estimator, we compute the divergence directly. This provides more accurate (and deterministic) computation than using the approximation, and is feasible in our case, given this paper focuses on low-dimensional problems.

\subsection{Sinusoidal Inverse Problem}

\paragraph{System Description.}
As described in the main text, we used a forward process defined by $y = f(x) + \epsilon$, where $f : \mathbb{R} \rightarrow \mathbb{R}$ is given by $f(x) = \tfrac{x}{2} + 0.7 \sin(5x)$ and $\epsilon \sim N(0, \sigma^2)$ with $\sigma = 0.2$. We then wish to compute the conditional probability distribution $p(x|y)$ of $x$ given some observation $y$.

\paragraph{Ground Truth Inverse Likelihood.} Although the inverse conditional likelihood $p(x|y)$ does not admit a closed form solution (as far as we are aware), it can be computed numerically by using Bayes' theorem:
\begin{align}
    p(x|y) &= \frac{p(y|x)p(x)}{p(y)}\\
        &= \frac{\mathcal{N}(y;f(x),0.2^2)\cdot \frac{1}{3}}{\int_{-1.5}^{1.5} p(y|x')p(x')dx'}\\
        &= \frac{\mathcal{N}(y;f(x),0.2^2)}{\int_{-1.5}^{1.5}\mathcal{N}(y;f(x'),0.2^2)dx'}.
\end{align}
We used the formula above to create the likelihood and NLL plots seen in Figure \ref{fig:inverse_sine_gt_distribution}.

\paragraph{Data Generation.}
We generated train datasets of varying sizes $N \in \{50, 100, 200, 500, 1000, 2000, 5000, 10000\}$. The inputs were sampled uniformly from the interval $[-1.5, 1.5]$. For all cases, we used an independent test dataset of size $1000$.

\paragraph{Model Architecture.}
We compared three model architectures MDN, CFM and a standard MSE-trained baseline. All approaches used the same backbone architecture of 5 dense layers of width $128$ with GeLU activation \cite{hendrycks2023gelu}. The MDN model used a total of $K=8$ mixture components. To robustly estimate the Negative Log-Likelihood (NLL) and quantify the variance due to initialization, we trained an ensemble of 12 models for each architecture-dataset pair.

\paragraph{Training.} All methods were trained using AdamW \cite{loshchilov2017adamW} for a total of $30,000$ iterations. The MDN model used a learning rate schedule of 100 linear warm-up steps to a peak LR of $5e-3$, followed by exponential decay of $0.9$ every $1,000$ steps. To achieve better results for CFM, which required smaller learning rates, we used $1,000$ linear warm-up steps to a peak LR of $1e-3$, followed by the same exponential decay schedule as MDN.

\begin{figure}[!ht]
    \centering
    \includegraphics[width=0.99\linewidth]{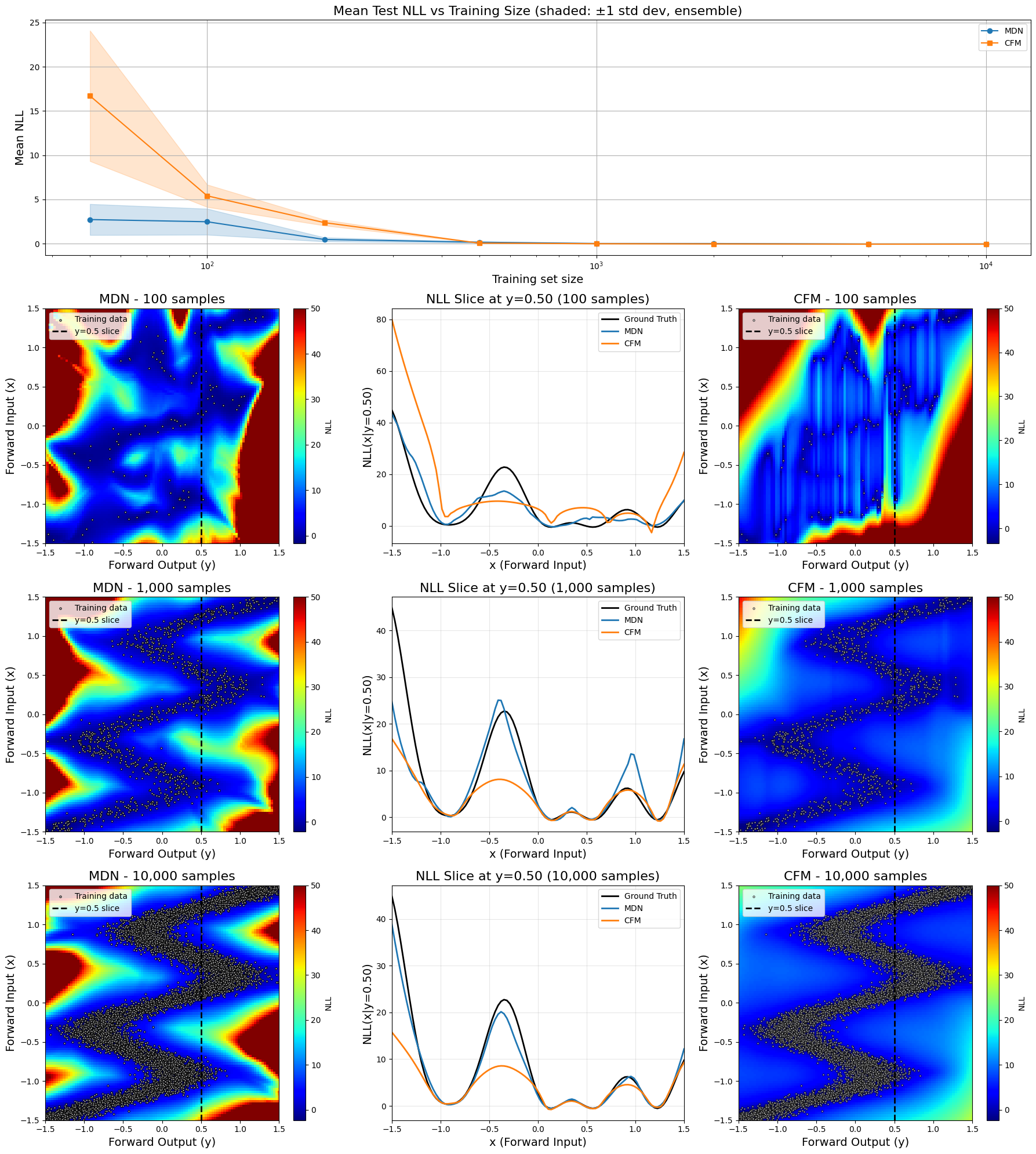}
    \caption{Comparison between MDN and CFM in the sinusoidal inverse problem. The top plot compares the test NLL (lower is better) of the two methods for differing sizes of the training set. Standard deviation bands correspond to results from 12 different starting seeds. The remaining plots display the NLL surface (averaged across seeds) on a subset of the experiments.}
    \label{fig:data-efficiency-inverse-sine_full}
\end{figure}

\paragraph{Results.}
As described in the main text, MDN significantly outperforms CFM in the low-data regime. Even in the high data setting, CFM continues to overestimate the likelihood of rare events. In Figure \ref{fig:data-efficiency-inverse-sine_full}, we extend the results from Figure \ref{fig:data-efficiency-inverse-sine} with additional NLL surface plots and slices of $-log(p(x \lvert y = 0.5))$, where the over smoothing tendency of CFM can be more readily seen. For the sake of completeness, we also report the values plotted on the top graph of Figures \ref{fig:data-efficiency-inverse-sine} \& \ref{fig:data-efficiency-inverse-sine_full} in Table \ref{tab:inverse_sine_data_efficiency}.

\begin{table}[ht]
    \centering
    \begin{tabular}{rcc}
        \toprule
        $N$ (samples) & MDN NLL $(\downarrow)$ & CFM NLL $(\downarrow)$\\
        \midrule
        50 & $2.727 \pm 1.753$ & $16.709 \pm 7.379$ \\
        100 & $2.478 \pm 1.469$ & $5.409 \pm 1.265$ \\
        200 & $0.723 \pm 0.462$ & $1.360 \pm 0.227$ \\
        500 & $0.235 \pm 0.074$ & $0.124 \pm 0.010$ \\
        1000 & $0.059 \pm 0.068$ & $-0.002 \pm 0.003$ \\
        2000 & $-0.017 \pm 0.006$ & $-0.036 \pm 0.002$ \\
        5000 & $-0.042 \pm 0.003$ & $-0.040 \pm 0.002$ \\
        10000 & $-0.040 \pm 0.003$ & $-0.049 \pm 0.001$ \\
        \bottomrule
    \end{tabular}
    \caption{Data efficiency comparison between MDN and CFM for the inverse sinusoid problem, shown in the top graph of Figure \ref{fig:data-efficiency-inverse-sine_full}.}
    \label{tab:inverse_sine_data_efficiency}
\end{table}

\begin{figure}[!ht]
    \centering
    \includegraphics[width=0.99\linewidth]{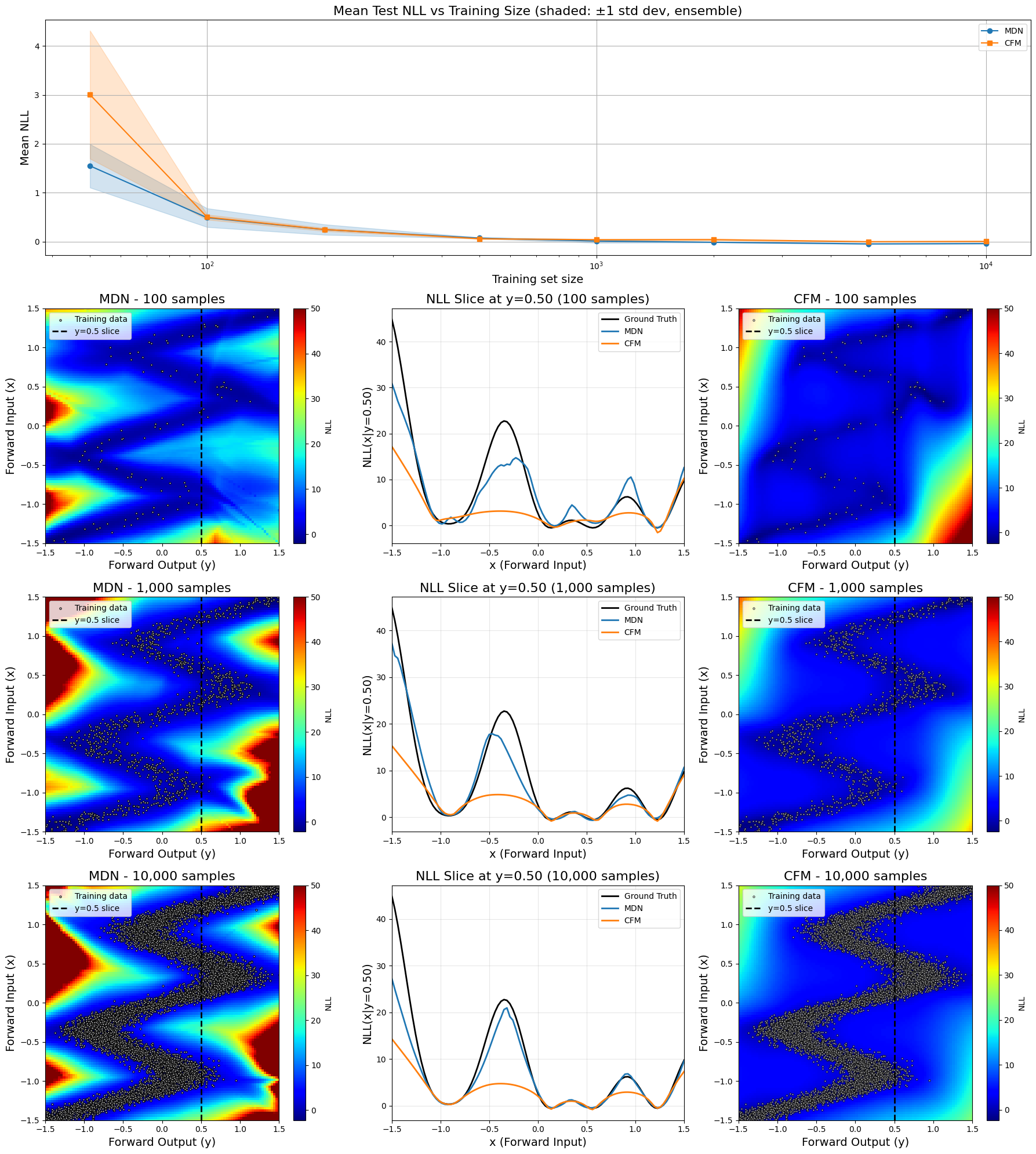}
    \caption{Data efficiency comparison between MDN and CFM approaches, analogous to Figure \ref{fig:data-efficiency-inverse-sine_full}, but using a smaller backbone network.}
    \label{fig:data-efficiency-inverse-sine_smaller_network}
\end{figure}

\paragraph{Network Size Ablation.}
As an ablation for the network size, we repeated the data-efficiency experiment shown in Figure $\ref{fig:data-efficiency-inverse-sine}$ with a smaller backbone network of only 3 layers of width $64$. Our findings in this scenario, which can be seen in Figure \ref{fig:data-efficiency-inverse-sine_smaller_network}, matched the results from the bigger network, with MDN outperforming CFM in the low-data regime. For this choice of network, we observed that even with larger datasets ($N=10,000$) MDN continued to outperform CFM in terms of NLL. Additionally, the ``bridge'' topological artifacts described in section \ref{sec:inductive-bias} are also seen, with CFM continuing to overestimate the likelihood of low-probability regions. For the sake of completeness, we also report the values plotted on the top graph of Figure \ref{fig:data-efficiency-inverse-sine_smaller_network} in Table \ref{tab:inverse_sine_data_efficiency_smaller_network}.

\begin{table}[ht]
    \centering
    \begin{tabular}{rcc}
        \toprule
        $N$ (samples) & MDN NLL $(\downarrow)$ & CFM NLL $(\downarrow)$\\
        \midrule
        50 & $1.548 \pm 0.445$ & $3.006 \pm 1.311$ \\
        100 & $0.487 \pm 0.192$ & $0.499 \pm 0.052$ \\
        200 & $0.244 \pm 0.106$ & $0.246 \pm 0.032$ \\
        500 & $0.071 \pm 0.011$ & $0.057 \pm 0.009$ \\
        1000 & $0.010 \pm 0.037$ & $0.037 \pm 0.015$ \\
        2000 & $-0.016 \pm 0.002$ & $0.038 \pm 0.021$ \\
        5000 & $-0.048 \pm 0.006$ & $-0.003 \pm 0.003$ \\
        10000 & $-0.043 \pm 0.003$ & $0.003 \pm 0.007$ \\
        \bottomrule
    \end{tabular}
    \caption{Data efficiency comparison between MDN and CFM for the inverse sinusoid problem using a smaller backbone network, shown in the top graph of Figure \ref{fig:data-efficiency-inverse-sine_smaller_network}.}
    \label{tab:inverse_sine_data_efficiency_smaller_network}
\end{table}

\begin{figure}[ht]
    \centering
    \includegraphics[width=0.49\textwidth]{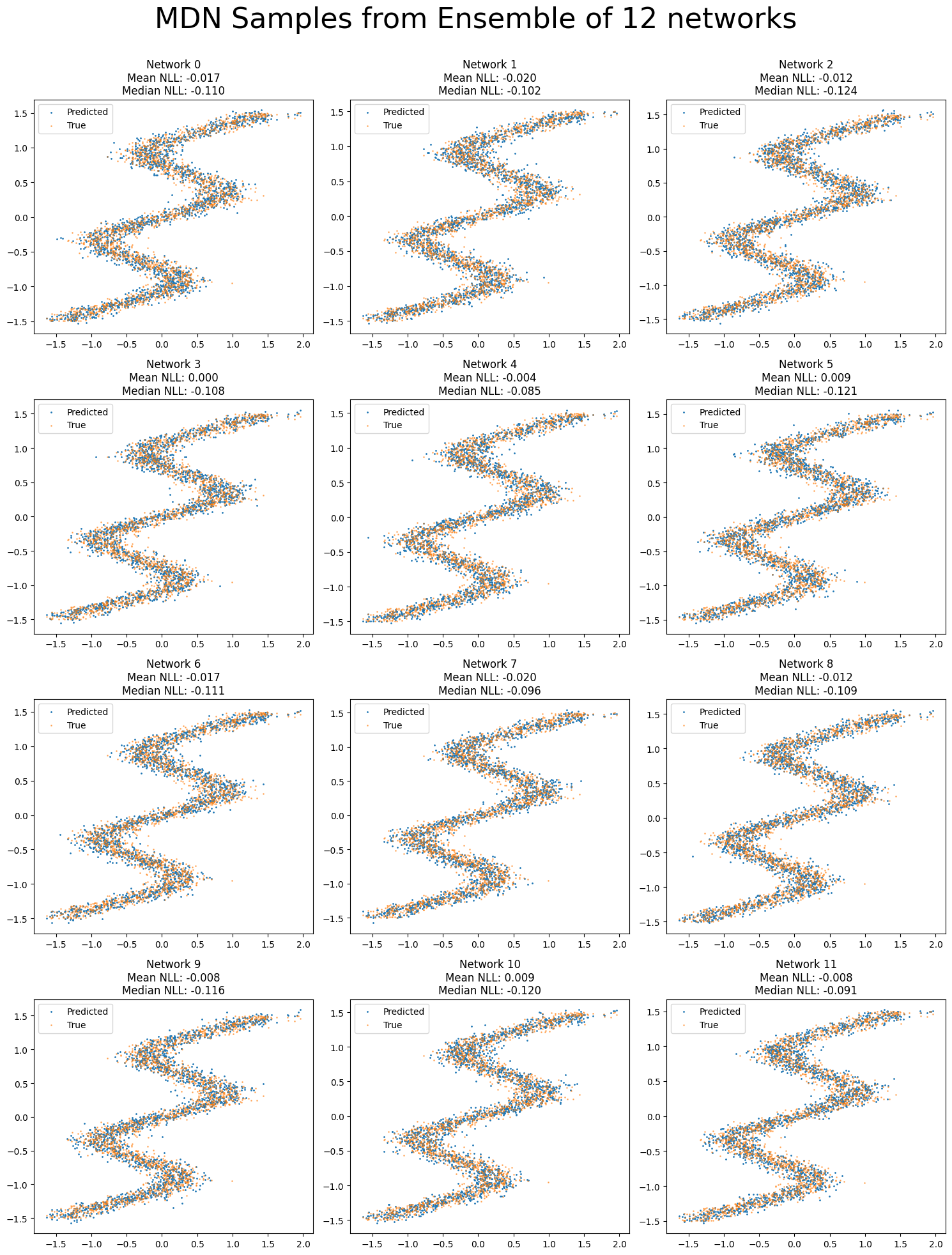}
    \vline
    \includegraphics[width=0.49\textwidth]{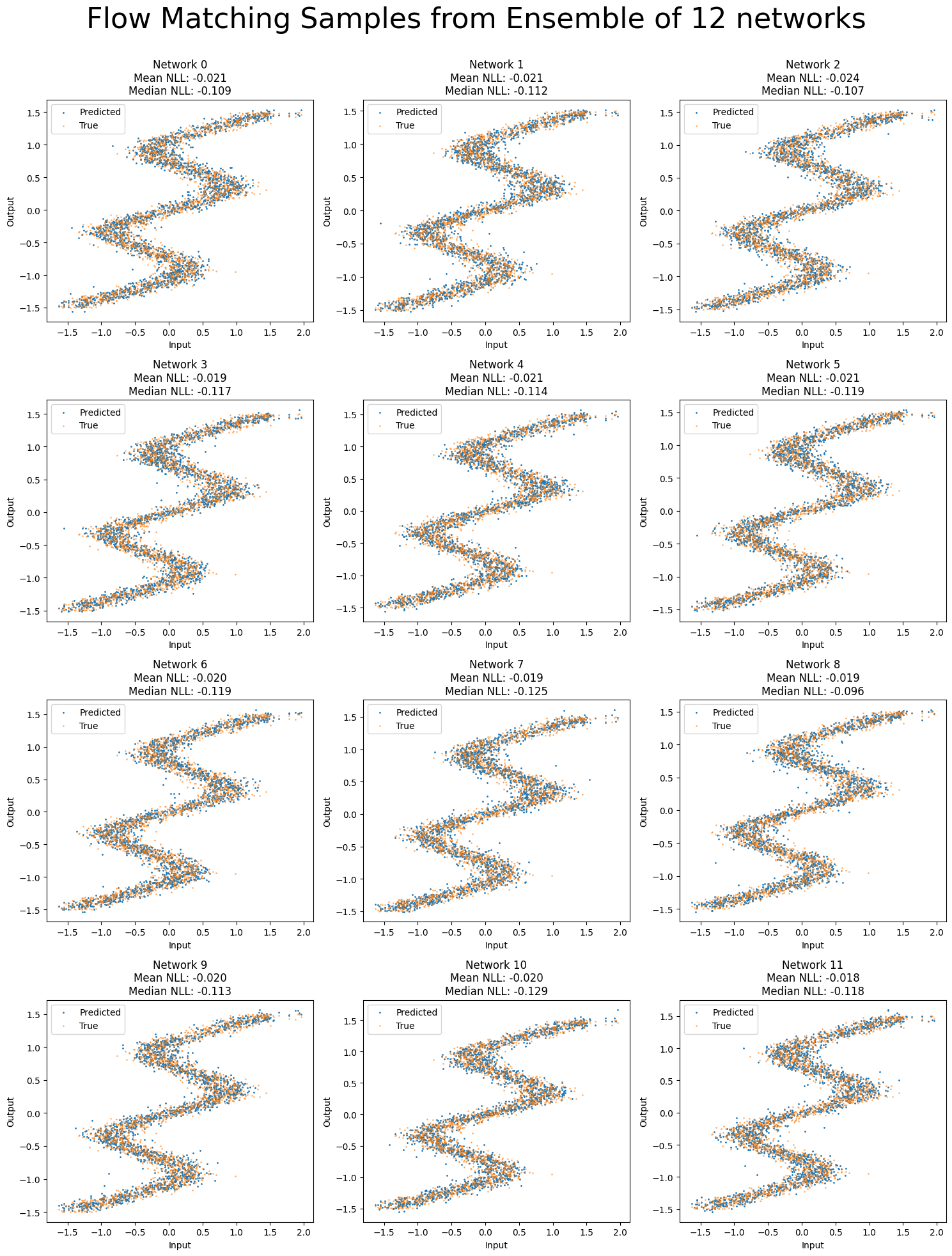}
    \caption{Representative samples from the conditional posterior distribution on the inverse sine problem using an ensemble of 12 networks. (left) MDN. (right) CFM.}
    \label{fig:samples_inverse_sine_mdn_and_cfm}
\end{figure}

\begin{figure}[ht]
    \centering
    \includegraphics[width=0.99\linewidth]{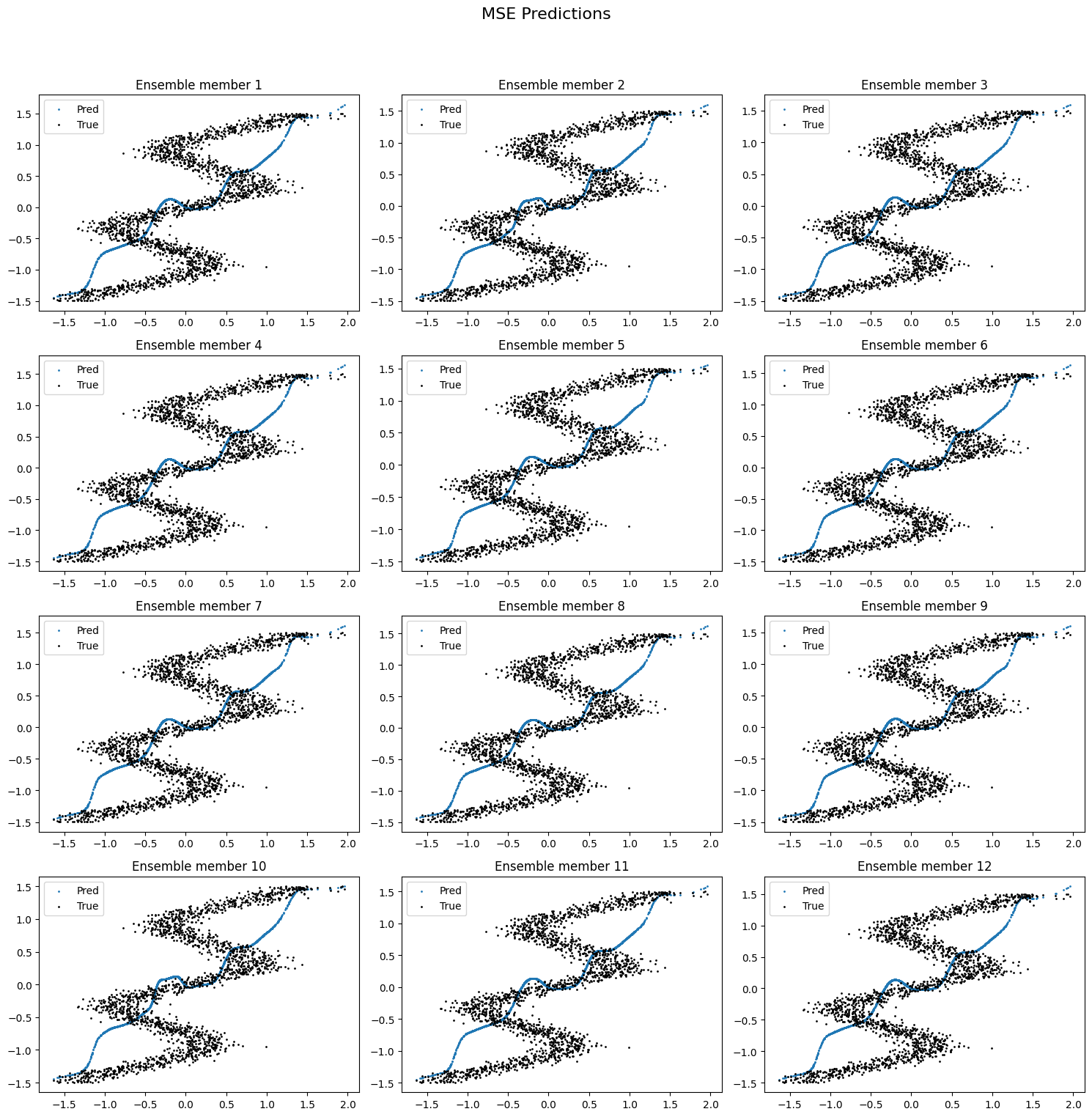}
    \caption{Point predictions for an MSE-trained ensemble of networks on the inverse sine problem. By predicting only the mean value, the network failed to capture the true structure of the dataset.}
    \label{fig:mse_inverse_sine}
\end{figure}

\paragraph{Failure of the Point Estimate.}
The MSE baseline performs catastrophically across all data regimes. As expected, it learns the conditional expectation $\mathbb{E}[x|y]$, which corresponds to the arithmetic mean of the conditional samples, as shown in Figure \ref{fig:mse_inverse_sine}. This mean often lies in a region of near zero probability density. This underscores the necessity of probabilistic modeling for inverse problems.

\clearpage
\subsection{Gravitational Attractor}

\paragraph{System Dynamics.}
We simulate the trajectory of a particle on a 2D plane under the gravitational influence of up to three static celestial bodies located at fixed positions $\mathbf{c}_1, \mathbf{c}_2, \mathbf{c}_3$. The equation of motion for the position vector $\mathbf{r}$ is given by:
\begin{equation}
    \ddot{\mathbf{r}} = -\sum_{j=1}^3 \frac{G M_j (\mathbf{r} - \mathbf{c}_j)}{||\mathbf{r} - \mathbf{c}_j||^2} - \gamma \dot{\mathbf{r}}^2,
\end{equation}
where $\gamma$ represents a damping coefficient (drag), chosen to be $\gamma=10$. We set the gravitational constant to $G=100$ and each body's mass to $M_j=1$. The positions of the attractors are evenly spaced on the unit circle and the particle's original position is sampled uniformly randomly on the disk with radius $0.2$ centered around the origin. Due to the damping term, the system is dissipative; every trajectory eventually loses energy and asymptotically converges to the location of one of the three bodies, which act as stable attractors. We also add a small isotropic observational noise that is normally distributed with mean 0 and standard deviation 0.05.

\begin{figure*}[ht]
    \centering
    \includegraphics[width=0.32\textwidth]{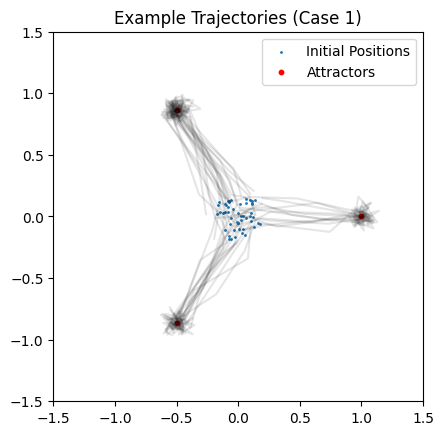}
    \vline
    \includegraphics[width=0.32\textwidth]{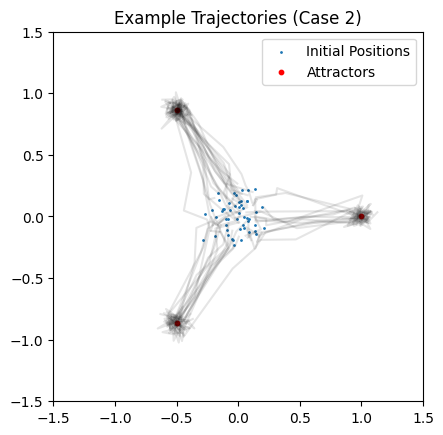}
    \vline
    \includegraphics[width=0.32\textwidth]{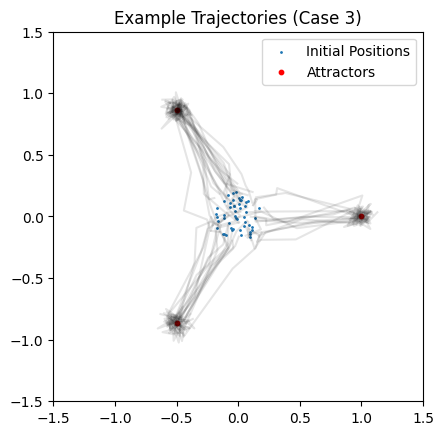}
    \caption{Example trajectories for the three different cases considered in the gravitational attractor example. Despite having qualitatively very similar trajectories, each system is fundamentally different from the others. MDNs, however, are able to provide insight into the problem's structure by learning the input dependence of trajectories in a human-interpretable way, as can be seen in Figure \ref{fig:attractor_interpretability}.}
    \label{fig:attractor_trajectory_examples}
\end{figure*}

\paragraph{Description of Cases Considered.}
As described in the main text, we consider three variations of this system. Example trajectories for each case can be seen in Figure \ref{fig:attractor_trajectory_examples}.
\begin{enumerate}
    \item \textbf{Partially Observed System Without Input Noise:} (many possibilities of the attractor position): In this scenario, for each trajectory a single attractor is picked at random, and it is the only one used to evolve the system. This becomes a problem of partial observability, where we need different world models to account for each possibility. In this case, the total number of possible ``worlds'' is equal to the number of possible attractors, and we expect each ``active'' mixture to be picked with no dependence on the initial position.
    \item \textbf{Consistent System With Input Noise}: In this scenario, the governing system is consistent (all three attractors are always present), but inputs in the data are corrupted with observational noise. This makes it so that points close to the decision boundary of which attractor pulls strongest may not be reliable. Therefore, we expect each mixture element to be more certain (closer to either 1 or 0) farther from this decision boundary, and fuzzier near the origin.
    \item \textbf{Consistent System Without Input Noise:} (MDN effectively functions as MoE): In this scenario, all three attractors are always presents, and inputs in the data are clean. In this setting, there may not seem to be much of an advantage to the MDN approach, but as we can see, the method still works, and instead functions as a Mixture of Experts (MoE): for a given input, we pick a specific mixture with probability close to 1, which specializes on points at a given slice of the plane.
\end{enumerate}

\paragraph{Data Generation.}
The ground-truth trajectories of the particles were computed using a simple Euler integration scheme from $t=0$ up to $t=1$, using a step size of $dt=0.0005$, which results in 2,000 time steps. For training and inference, we uniformly subsample the trajectories to only 11 time steps $t\in\{0.0, 0.1, 0.2, \dots, 1.0\}$.

\paragraph{Model Architectures.}
Both MDN and CFM used the same backbone architecture of four hidden layers of dimension 128 and GeLU activation, with the MDN using 10 mixture components. To control for variability in network initialization, we trained an ensemble of 12 networks for each method.

\paragraph{Training.}
Both MDN and CFM were trained for 30,000 steps using AdamW with weight decay of $0.01$ and batch size 64. The learning rate followed a schedule of 500 linear warm-up steps to a peak LR of 5e-4, followed by an exponential decay rate of $0.9$ every 1,000 steps, plateauing at 5e-4. We additionally used adaptive gradient clipping \cite{adaptive_grad_clip-2021} with a rate of $0.01$.

\paragraph{Comparison with Implicit Models.}
We performed the same experiment using a Conditional Flow Matching model. MDN consistently outperforms CFM in terms of test NLL, as can be seen in Table \ref{tab:gravitational_attractor_test_nll}. While the CFM also accurately predicted the marginal distribution of outcomes (generating samples clustered around the three attractors), it provided no direct mechanism to visualize the basin structure. This result substantiates the argument in Section \ref{sec:inductive-bias}: the parametric structure of the MDN is not merely a computational convenience but a semantic feature that aligns with the discrete nature of multi-stable physical systems, offering ``white-box'' insight into the phase topology that implicit models obfuscate.

\begin{table}[ht]
    \centering
    \begin{tabular}{rcc}
        \toprule
         & MDN NLL $(\downarrow)$ & CFM NLL $(\downarrow)$\\
        \midrule
        Case 1 & $\mathbf{-32.596 \pm 0.059}$ & $-30.583 \pm 0.096$ \\
        Case 2 & $\mathbf{-28.445 \pm 0.184}$ & $-27.504 \pm 0.083$ \\
        Case 3 & $\mathbf{-32.840 \pm 0.088}$ & $-30.718 \pm 0.177$ \\
        \bottomrule
    \end{tabular}
    \caption{Test NLL for MDN and CFM in each of the cases considered in the gravitational attractor problem.}
    \label{tab:gravitational_attractor_test_nll}
\end{table}

\clearpage
\subsection{ODE Saddle Node Bifurcation}

Bifurcations occur in dynamical systems when a small change in a dynamical coefficient leads to drastic changes in the behavior of the system. If this coefficient is unknown and potentially different across different dataset members (in the case of under-specified systems), accurate UQ becomes essential in order to determine possible trajectories for the quantities of interest. This has been the focus of recent papers in SciML such as \cite{hendriks2025equivariantflowmatchingsymmetrybreaking, hendriks2025wallpaper, psarellis2025active}. We showcase an example of such a system in ODEs, where multimodal distributions are necessary to capture the underlying uncertainty and yield useful predictions.

\paragraph{System Description.}
The saddle-node bifurcation is a fundamental mechanism by which fixed points are created and destroyed in dynamical systems. We consider the ODE:
\begin{equation}\label{eq:saddle_node_app}
\frac{dx}{dt} = r + x^2,
\end{equation}
where $r \in \mathbb{R}$ is the bifurcation parameter. The qualitative behavior of this system depends on the sign of $r$:

\begin{itemize}
    \item \textbf{Stable regime ($r < 0$):} Two fixed points exist at $x^* = \pm\sqrt{-r}$. The fixed point at $x^* = -\sqrt{-r}$ is stable (attracting), while $x^* = +\sqrt{-r}$ is unstable (repelling). Trajectories starting below the unstable fixed point ($x_0<-\sqrt{-r}$) converge to the stable equilibrium, while ones starting above it ($x_0>-\sqrt{-r}$) diverge to infinity. 
    
    \item \textbf{Critical point ($r = 0$):} The two fixed points collide and annihilate at the origin in a saddle-node bifurcation, where a semi-stable fixed point exists.
    
    \item \textbf{Unstable regime ($r > 0$):} No fixed points exist and all trajectories diverge to infinity. However, a remnant of the bifurcation persists in the form of critical slowing down, trajectories exhibit a deceleration as they pass through the bottleneck region near $x \approx 0$.
\end{itemize}

\paragraph{Data Generation.}
We generate training data by numerically integrating equation (\ref{eq:saddle_node_app}) using a forward Euler scheme with $\Delta t = 0.001$ over a total time of $T = 5.0$. Trajectories are then subsampled to 20 time steps. Initial conditions $x_0$ are sampled uniformly from $[-2.0, 2.0]$, and the bifurcation parameter $r$ is sampled from a discrete set of 5 values $r\in\{-1.50, -0.75, 0.00, 0.75, 1.50\}$ uniformly spaced in $[-1.5, 1.5]$. Observation noise with standard deviation $\sigma = 0.1$ is added to the trajectories. Crucially, $r$ is treated as a hidden variable: the model only receives $x_0$ as input and must predict the full discretized trajectory $x(t_1), x(t_2), \dots, x(T)$. Additionally, for trajectories that tend to infinity, we clip positions at $x=10$ to avoid the need of representing arbitrarily large numbers.

The test set contains $N_{\text{test}} = 2000$ samples, held fixed across all experiments.

\paragraph{Model Architectures.}
Both MDN and CFM models used a backbone architecture consisting of a five-layer MLP with 256 units per layer and GeLU activations \cite{hendrycks2023gelu}. To capture seed variability, both methods were trained using an ensemble of 12 independent initializations. The MDN uses $K=15$ mixture components.

\paragraph{Training.}
Training for both models is conducted over 50,000 iterations with a batch size of 128 using the AdamW optimizer and $10^{-5}$ weight decay. The learning rate follows a schedule with a 2,000-step linear warmup to a peak of $5 \times 10^{-4}$, followed by continuous exponential decay of $0.9$ every 2,000 steps. We also used adaptive gradient clipping \cite{adaptive_grad_clip-2021} with a clip rate of $0.1$.

\paragraph{Sample Efficiency Results.}
Table~\ref{tab:saddle_node_efficiency} reports the test NLL (mean $\pm$ std across 12 ensemble members) for varying training set sizes. MDN achieves better NLL than CFM in 50--100 samples, while CFM requires approximately 200+ samples to produce reasonably better estimates. At $N \geq 5000$, both methods converge to similar performance.

\begin{table}[ht]
\centering
\caption{Sample efficiency comparison for saddle-node bifurcation system. NLL values reported as mean $\pm$ std across 12 ensemble members. Lower is better.}
\label{tab:saddle_node_efficiency}
\begin{tabular}{rcc}
\toprule
$N$ (samples) & MDN NLL $(\downarrow)$& CFM NLL $(\downarrow)$\\
\midrule
50 & $87.3 \pm 7.2$ & $665.9 \pm 139.9$ \\
100 & $77.4 \pm 3.5$ & $146.0 \pm 21.9$ \\
200 & $50.6 \pm 13.1$ & $16.1 \pm 2.0$ \\
500 & $3.4 \pm 1.4$ & $-9.5 \pm 0.2$ \\
1000 & $-5.5 \pm 1.8$ & $-12.8 \pm 0.1$ \\
2000 & $-10.9 \pm 1.3$ & $-13.5 \pm 0.1$ \\
5000 & $-13.4 \pm 0.9$ & $-13.9 \pm 0.1$ \\
10000 & $-13.9 \pm 0.3$ & $-14.0 \pm 0.1$ \\
\bottomrule
\end{tabular}
\end{table}

\begin{figure}[ht]
    \centering
    \includegraphics[width=0.9\linewidth]{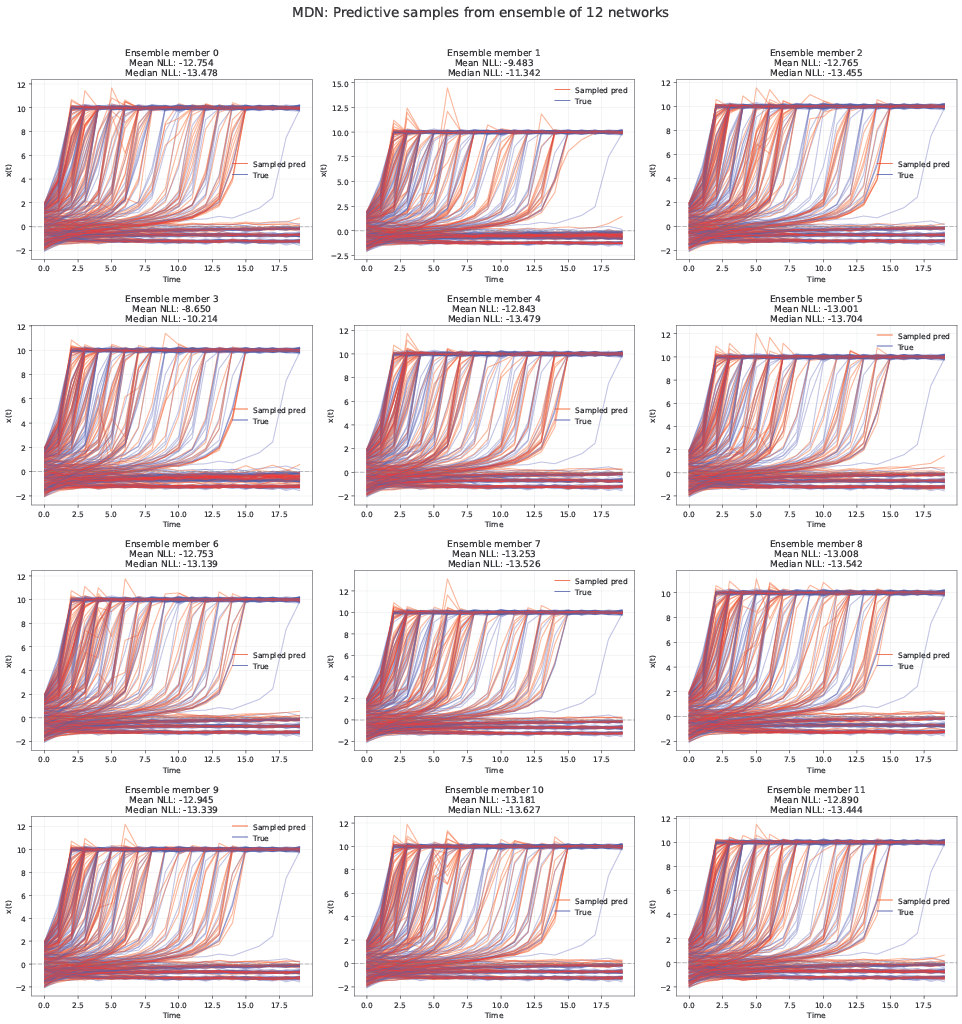}
    \caption{MDN ensemble predictions on the saddle-node bifurcation system. Each subplot shows samples from a single ensemble member's predicted distribution for the same set of initial conditions. The ensembles exhibits consistent multimodal structure with clear separation between convergent and divergent trajectories across all 12 members.}
    \label{fig:mdn_ensemble_saddle_node}
\end{figure}

\begin{figure}[ht]
    \centering
    \includegraphics[width=0.9\linewidth]{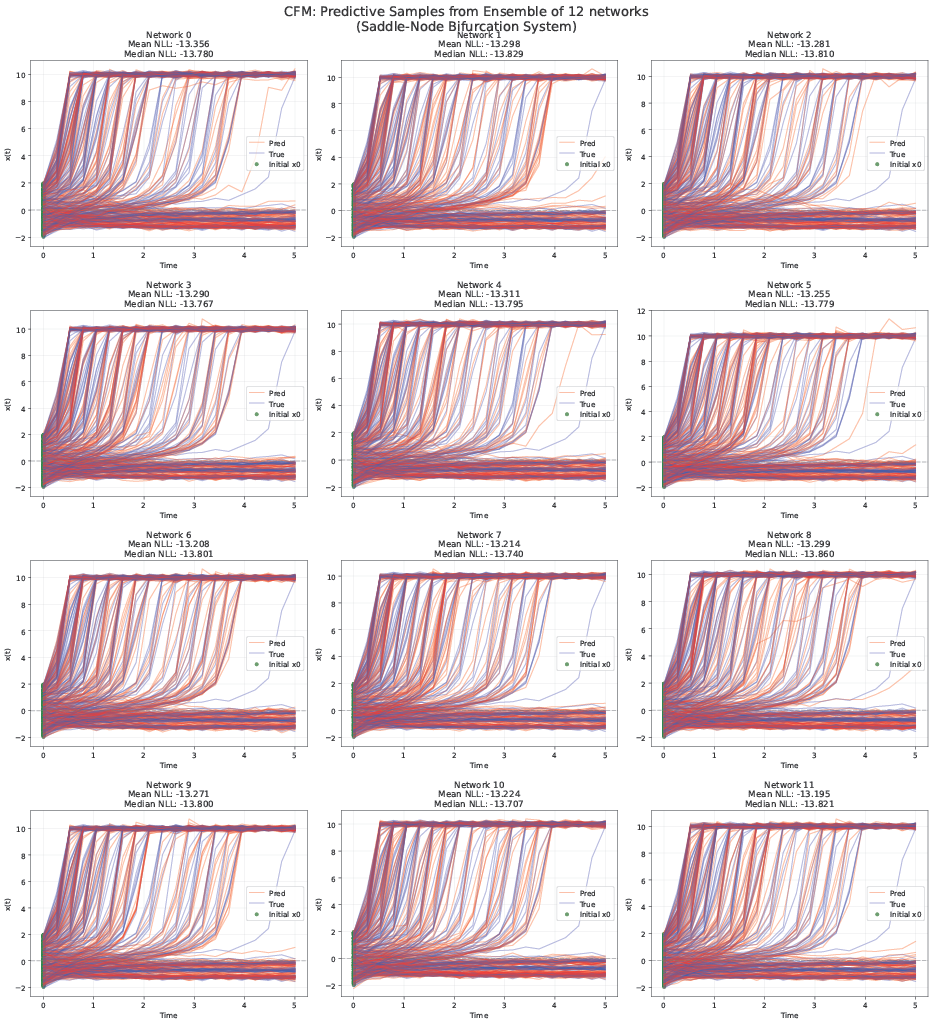}
    \caption{CFM ensemble predictions on the saddle-node bifurcation system. Each subplot shows samples from a single ensemble member's predicted distribution for the same set of initial conditions. The CFM ensemble equally captures bimodality in the saddle-node dynamics.}
    \label{fig:cfm_ensemble_saddle_node}
\end{figure}

\paragraph{Successful Multimodal Predictions: MDN vs.\ CFM.}
Figures~\ref{fig:mdn_ensemble_saddle_node} and~\ref{fig:cfm_ensemble_saddle_node} compare the predictions from all 12 ensemble members for both MDN and CFM. Both methods successfully capture the multimodal nature of the saddle-node bifurcation, correctly identifying the distinct convergent ($r < 0$) and divergent ($r > 0$) dynamical regimes. The MDN ensemble (Figure~\ref{fig:mdn_ensemble_saddle_node}) exhibits consistent multimodal behavior with well-separated modes across all members, while the CFM ensemble (Figure~\ref{fig:cfm_ensemble_saddle_node}) also captures the bimodal structure.

\begin{figure}[ht]
    \centering
    \includegraphics[width=0.9\linewidth]{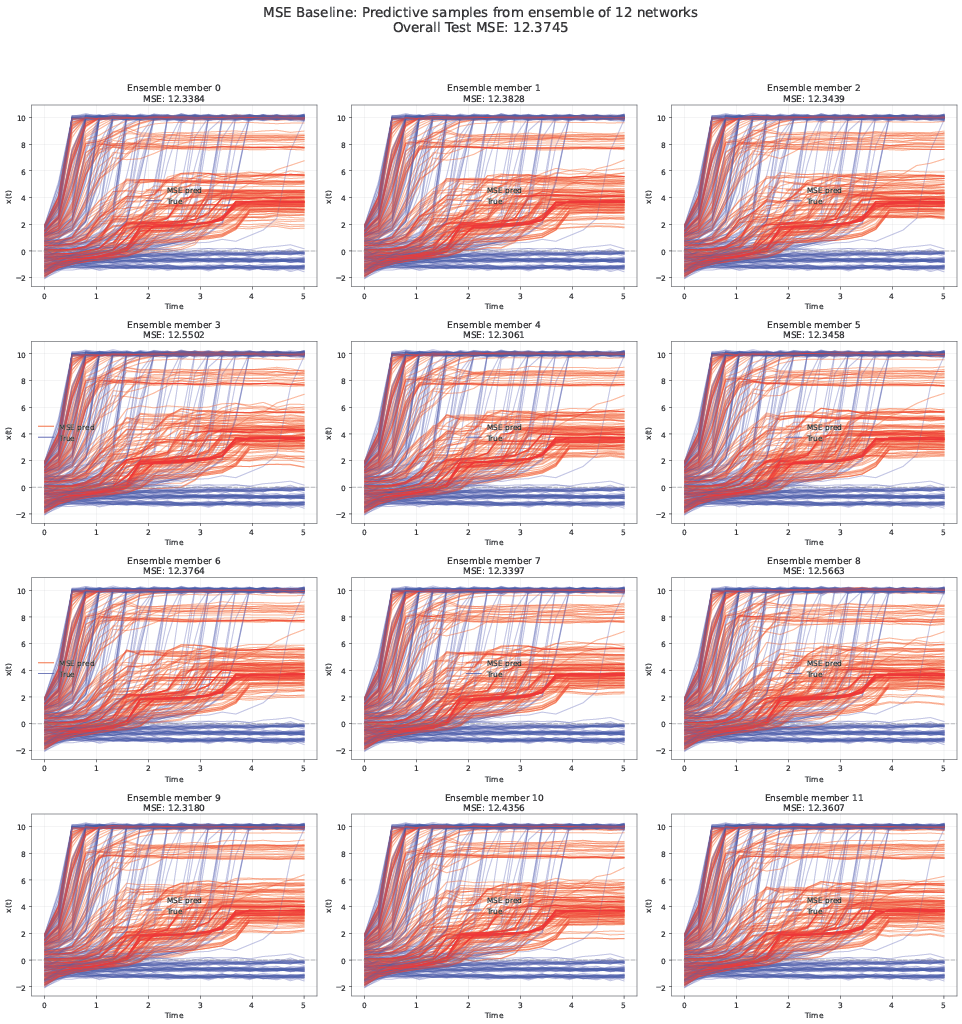}
    \caption{Failure mode of MSE-trained ensemble on the saddle-node bifurcation system. Each subplot shows the deterministic prediction from a single ensemble member. Unlike MDN and CFM, all ensemble members collapse to similar predictions near the conditional mean, which lies in a region of low probability density between the two physical regimes. The predicted trajectories correspond to an average of mutually exclusive outcomes and do not represent any physically realizable state.}
    \label{fig:mse_ensemble_saddle_node}
\end{figure}

\paragraph{Failure Mode: MSE Point Estimates.}
Figure~\ref{fig:mse_ensemble_saddle_node} illustrates the fundamental failure mode of point-estimate networks trained with Mean Squared Error loss. All 12 ensemble members predict trajectories that cluster around the conditional mean, which lies between the two physically realizable outcomes. This ``averaging'' behavior produces predictions that correspond to neither the convergent nor divergent regime, effectively predicting a trajectory that would never be observed in the actual dynamical system. The tight clustering of MSE predictions around the mean, rather than around the true modes, demonstrates the inadequacy of point-estimate methods for problems with multimodal uncertainty.

\clearpage
\subsection{Lorenz System}

\paragraph{System Description.}
We consider the Lorenz-63 system \cite{lorenz1963deterministic} with 3D position $(x,y,z)$ evolved as:
\begin{align}
    \dot{x} &= \sigma(y-x), \\
    \dot{y} &= x(\rho-z)-y, \\
    \dot{z} &= xy - \beta z,
\end{align}
with standard parameters $\sigma=10, \beta=8/3, \rho=28$. This system is characterized by a positive leading Lyapunov exponent $\lambda_1 \approx 0.906$, implying that any infinitesimal error in the initial state grows exponentially as $\|\delta(t)\| \approx \|\delta_0\|e^{\lambda_1 t}$. For long prediction horizons ($t \gg 1/\lambda_1$), point-wise prediction is impossible. In the context of machine learning, a trustworthy prediction for this problem is therefore not a single trajectory, but the probability density on possible trajectories.

\paragraph{Data Generation.}
The trajectories used to train the models were obtained using the Diffrax package \cite{kidger2021diffrax} with the Dormand-Prince's (DOPRI) 5/4 solver \cite{dormand1980family} and a time step of $dt=0.001$ from time $t=0$ to $t=10$, resulting in $10,000$ observations per trajectory. For training and inference, we uniformly subsample trajectories to contain only $500$ observations per trajectory, yielding a time interval of $\Delta t = 0.02$. The initial positions were randomly sampled from a normal distribution with unit variance and mean $(0, 0, 24.5)\in\R^3$, to place particles near the trajectory manifold of the dynamical system and minimize burn-in of the initial simulation steps. We also add a small isotropic observational noise that is normally distributed with mean 0 and standard deviation 0.2.

\paragraph{Model Architectures.}
The MSE and plain MDN models used the same backbone architecture of 5 hidden layers of width 128 and GeLU activation, with MDN using 9 mixtures. This resulted in 66,947 and 74,687 total parameters for MSE and MDN, respectively. The RNN-MDN consisted of a single GRU layer \cite{chung2014gru} with hidden dimension 128 and 8 mixture components, resulting in 58,040 total parameters. Notably, RNN-MDN achieves superior performance with the fewest parameters (58K vs.\ 67K for MSE), indicating that the benefit stems from the probabilistic output structure rather than increased capacity.

\begin{figure}[ht]
    \centering
    \includegraphics[width=0.49\textwidth]{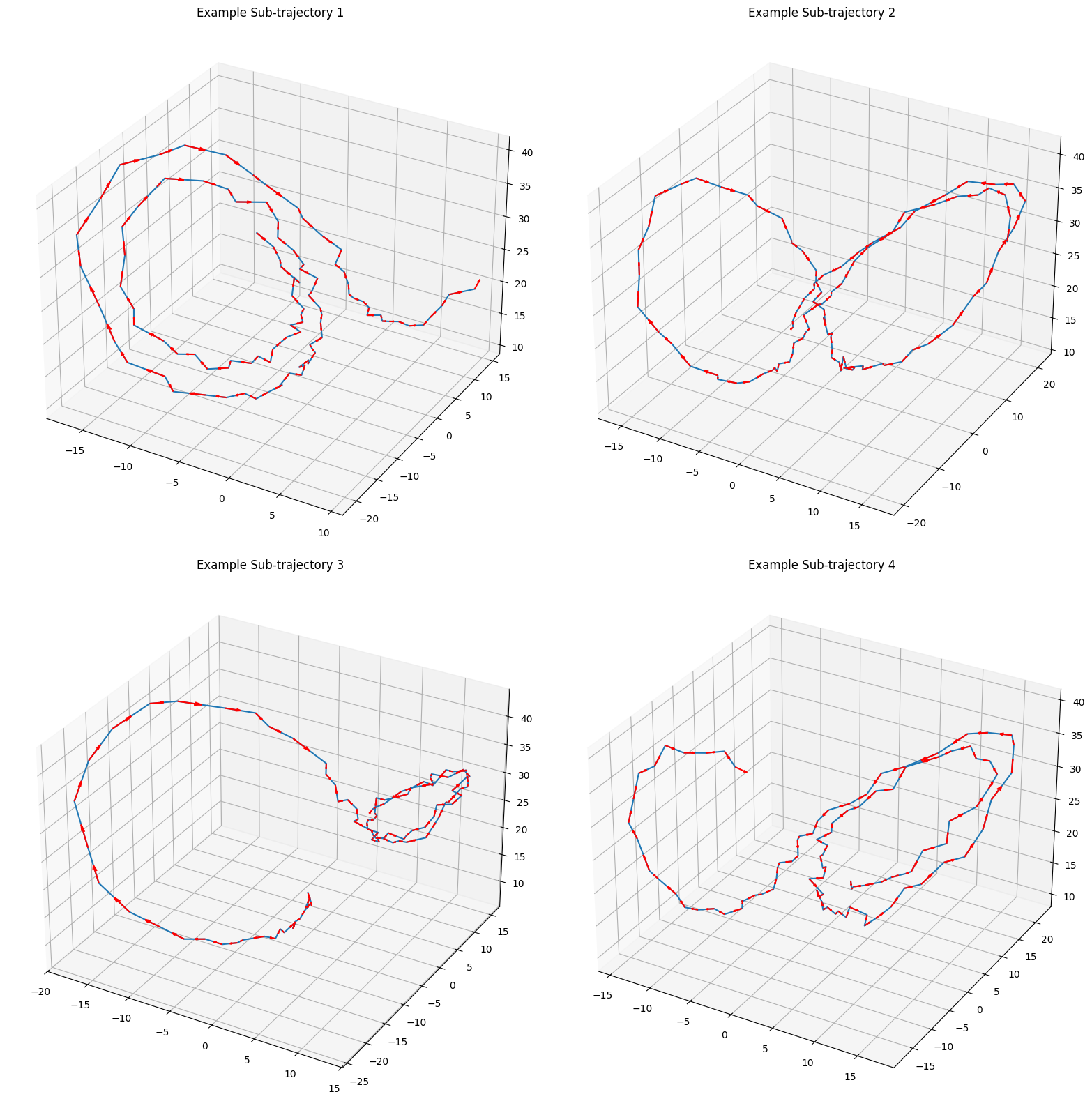}
    \caption{Example sub-trajectories used to train the RNN-MDN model for the Lorenz system.}
    \label{fig:rnn-mdn-example-subtrajectories}
\end{figure}

\paragraph{Training.}
All three models were trained using AdamW with weight decay of 0.1 for 30,000 iterations. The learning rate followed a linear warm-up to peak LR of $1e-3$ for 1,000 iterations, followed by an exponential decay rate of 0.9 every 1,000 steps. MSE and MDN models were trained using a batch size of 256, while the RNN-MDN used a batch size of 5 sub-trajectories of length 100. Four representative examples of training sub-trajectories can be seen in Figure \ref{fig:rnn-mdn-example-subtrajectories}.

\paragraph{Results}
We trained autoregressive models to predict the state transition $\Delta x_t := x_{t+\Delta t}-x_t$. We then performed rollouts to observe the long-term behavior of the generated trajectories. We compared a standard network trained with MSE against a plain MDN and a Recurrent Mixture Density Network (RNN-MDN) \cite{ha2018WorldModels, ha2017SketchRNN}. The MSE-trained model failed to capture the true properties of the trajectory in the long run, while the RNN-MDN successfully captured the aleatoric uncertainty. The RNN-MDN trajectory stays on the manifold of the true Lorenz attractor indefinitely.

This behavior, can be qualitatively seen in the trajectories plotted in Figure \ref{fig:lorenz-trajectories}: while the MSE model produces orbits that are too large because it averages possible changes in position over the attractor, the MDN model is better able to capture the true variability of the system, but over long time horizons creates overly noisy trajectories; finally, the RNN-MDN model is the only one that successfully captures the geometry and topology of the solution manifold, correctly recovering the two ``holes'' near each attractor. Our quantitative analysis comparing the Maximum Mean Discrepancy (MMD) between ground truth and predictions of each method also showcase RNN-MDN as the best performing method, as can be seen in Table \ref{tab:lorenz-mmd} and its graphical representation in Figure \ref{fig:lorenz-mmd-bar-graph}.

\begin{figure}[ht]
    \centering
    \includegraphics[width=0.75\linewidth]{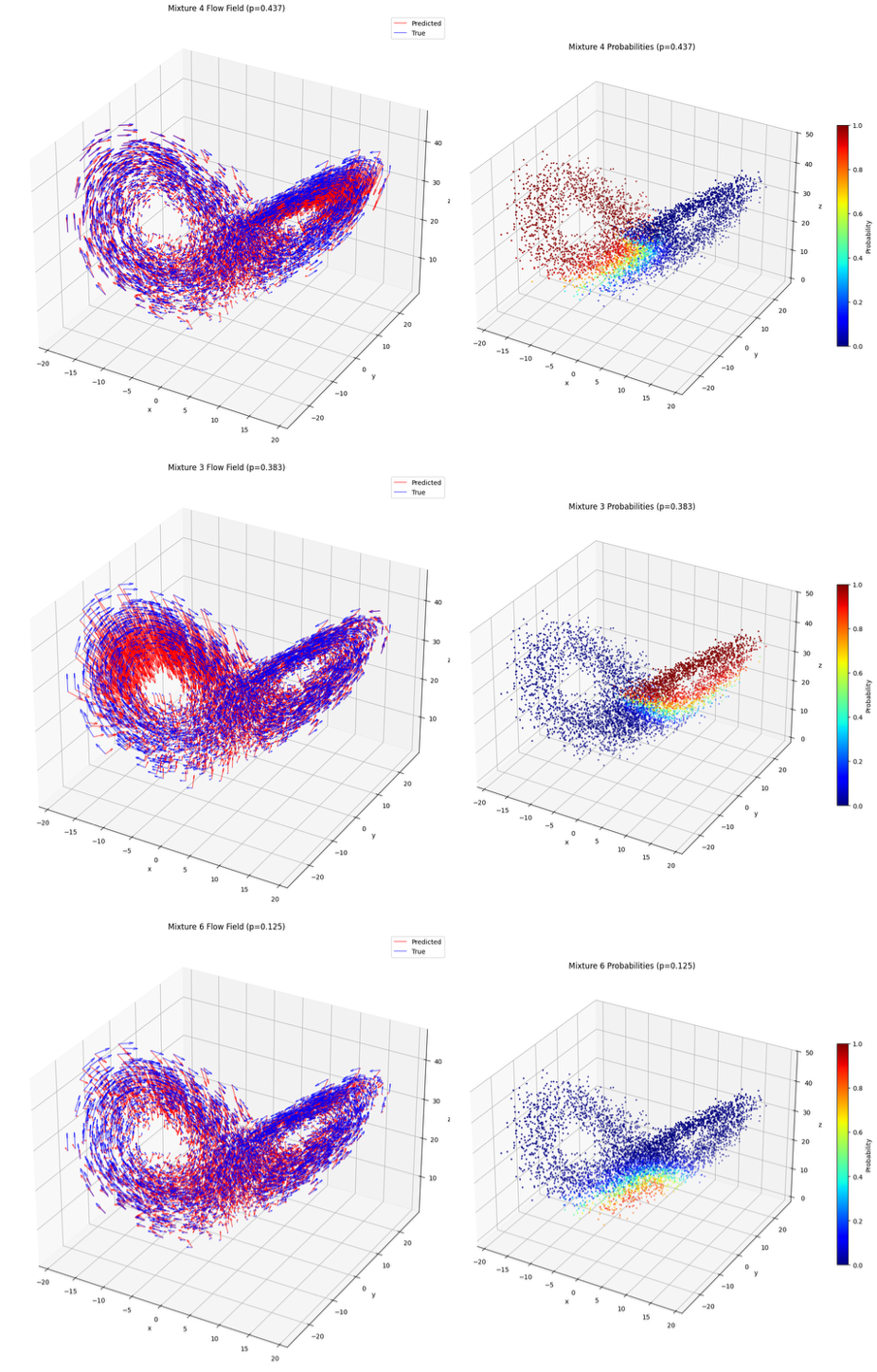}
    \caption{Top three mixture components by marginal probability for RNN-MDN in the Lorenz problem. The left column shows the predicted and true displacement fields, while the right column shows a color plot of the mixture weights based on the input.}
    \label{fig:rnn-mdn-mixture-decomposition}
\end{figure}

\paragraph{Interpretability}
The RNN-MDN is able to learn a meaningful partition of the space in an unsupervised manner, learning to dedicate distinct mixture components to each of the two ``lobes'' of the attractor. This behavior can be seen in Figure \ref{fig:rnn-mdn-mixture-decomposition}, where the network successfully learns a meaningful decomposition of the space by separating the two-lobbed structure of the problem along with an ambiguous region in the intersection of the two structures.

\subsection{Maximum Mean Discrepancy (MMD) Computation}

Given the chaotic nature of the Lorenz system, point-for-point error estimates such as MSE or conditional likelihood do not yield useful information regarding the realism of predicted paths beyond very short time horizons. This problem is heightened by the fact that our system also includes inherent stochasticity due to observational noise. Instead, we compute statistical properties of the ground truth trajectories and compare them to the ones of our prediction models.

\begin{figure}[ht]
    \centering
    \includegraphics[width=0.99\linewidth]{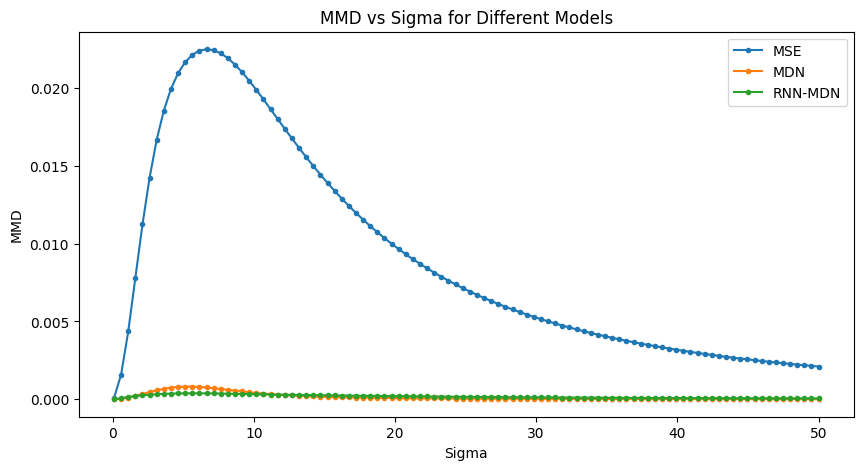}
    \caption{MMD values for different method with varying sigma values in $\sigma\in[0.1, 50]$ for the Lorenz problem.}
    \label{fig:lorenz-mmd-vs-sigma}
\end{figure}

\begin{figure}[ht]
    \centering
    \includegraphics[width=0.45\textwidth]{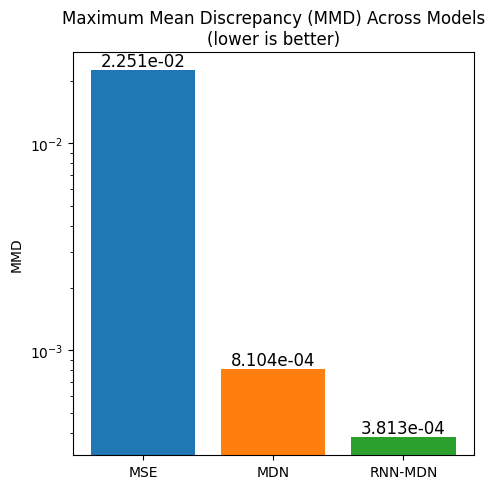}
    \caption{Maximum MMD values for different methods for the Lorenz problem, as reported in Table \ref{tab:lorenz-mmd}.}
    \label{fig:lorenz-mmd-bar-graph}
\end{figure}

One common approach in such a scenario is to compute the empirical Maximum Mean Discrepancy (MMD) between the point clouds of positions predicted by different methods. For two point clouds $X=(x_1, \dots x_{n_1})$ and $Y=(y_1, \dots y_{n_1})$ with points in $\R^d$, this is done by computing map $\phi:\R^d\to \mathcal{F}$ from $\R^d$ to some feature space $\mathcal{F}$. The MMD between the two datasets is then computed as the distance between feature means:
\begin{equation}
    \left|\left|\frac{1}{n_1}\sum_{i=1}^{n_1}\phi(x_i) - \frac{1}{n_2}\sum_{i=1}^{n_2}\phi(y_i) \right|\right|_\mathcal{F}.
\end{equation}
Using the kernel trick, this can be equivalently computed with an appropriate kernel $k:\R^d\times\R^d\to \R$ by
\begin{align*}    
    MMD_k(X, Y) &= \left(\frac{1}{n_1(n_1-1)}\sum_{i=1}^{n_1}\sum_{j\neq i} k(x_i, x_j) \right)\\
        &\quad - \left(\frac{1}{n_1n_2}\sum_{i=1}^{n_1}\sum_{j=1}^{n_2} k(x_i, y_j) \right) \\
        &\quad + \left(\frac{1}{n_2(n_2-1)}\sum_{i=1}^{n_2}\sum_{j\neq i} k(y_i, y_j) \right).
\end{align*}
Since the choice of kernel can influence the properties that this MMD quantity captures, it is common to take the maximum over a family of kernels \cite{seidman2023vano}. For this work, we consider the RBF family of kernels with a choice of scale parameter $\sigma\in[0.1, 50]$. This maximum for each method is the bar graph of Figure \ref{fig:lorenz-mmd-bar-graph}, while full dependency on sigma values can be seen in Figure \ref{fig:lorenz-mmd-vs-sigma}.

